\documentclass[11pt]{article}

\usepackage[preprint]{acl}

\usepackage{times}
\usepackage{latexsym}

\usepackage[T1]{fontenc}

\usepackage[utf8]{inputenc}

\usepackage{microtype}

\usepackage{inconsolata}

\usepackage{graphicx}
\graphicspath{{img/}}

\usepackage{booktabs}
\usepackage{multicol}
\usepackage{multirow}
\usepackage{array}
\usepackage{longtable}
\usepackage{caption}

\usepackage{tipa}
\usepackage{amsmath}
\usepackage{amssymb}

\title{Phoneme- and Word-Level Metrics Using Self-Supervised Speech Representations for Forced Alignment Evaluation}

\author{
  \textbf{V.S.D.S. Mahesh Akavarapu\textsuperscript{1}},
  \textbf{Michael Daniel\textsuperscript{2}},
  \textbf{Gerhard J\"{a}ger\textsuperscript{1}}
\\
  \textsuperscript{1}University of T\"{u}bingen,
  \textsuperscript{2}University of Jena
\\
\texttt{mahesh.akavarapu@uni-tuebingen.de, misha.daniel@gmail.com,}\\
\texttt{gerhard.jaeger@uni-tuebingen.de} \\
}

\begin{document}
\maketitle
\begin{abstract}
Forced alignment evaluation typically requires manually annotated timestamps, limiting large-scale and multilingual analysis. We introduce two corpus-level metrics based on self-supervised (SSL) speech representations for reference-free forced alignment evaluation: Phoneme-Cluster Mutual Information (PCMI) and Word Acoustic Consistency Score (WACS). PCMI measures agreement between aligned phoneme labels and clusters induced from SSL-speech representations, while WACS measures consistency of repeated word realizations using dynamic time warping similarity between word representation sequences. Using both random and systematic perturbations, we show that PCMI and WACS degrade consistently under alignment perturbations. We further analyze the metrics across multiple alignment systems on 85 languages from FLEURS, validate them against manually annotated alignments from 45 languages in DoReCo, and evaluate them on two phonologically complex low-resource languages. The metrics effectively separate high- and low-quality alignments and correlate strongly with timestamp-based alignment quality measures. Our results demonstrate that SSL-speech representations enable scalable, reference-free forced alignment evaluation. The metrics are available as an open-source Python package at \url{https://github.com/mahesh-ak/forced-aligner-metrics}.
\end{abstract}

\section{Introduction}
\label{sec:intro}

Forced alignment maps speech waveforms to transcript units such as words or phonemes and finds applications in subtitle generation, speech corpus segmentation, acoustic-phonetic analysis, and speech synthesis. Recent years have seen increasing adoption of transformer-based speech models for alignment \citep{bain2023whisperx, rastorgueva2023nemo, shi2026qwen3}. However, evaluation of forced alignment systems still largely depends on manually annotated timestamps, which are expensive to obtain and predominantly available for English. Consequently, aligners are commonly evaluated either solely for English or against outputs of existing aligners such as Montreal Forced Aligner (MFA) \citep{mcauliffe2017montreal}. This substantially limits progress toward multilingual forced alignment.

\begin{table}[t]
\centering
\resizebox{\columnwidth}{!}{%
\begin{tabular}{llccc}
\toprule
\textbf{Emb.} & \textbf{Alignment model} & \textbf{\#Langs} & \textbf{PCMI} $\uparrow$ & \textbf{WACS} $\uparrow$ \\
\midrule
\multirow{5}{*}{MMS} & Baseline & 85 & 0.09 $\pm$ 0.02 & 0.02 $\pm$ 0.01 \\
 & MFA & 82 & 0.25 $\pm$ 0.11 & 0.14 $\pm$ 0.08 \\
 & MMS-300m-IPA* & 85 & 0.24 $\pm$ 0.03 & 0.19 $\pm$ 0.02 \\
 & Wav2Vec2-IPA* & 85 & 0.24 $\pm$ 0.03 & 0.19 $\pm$ 0.03 \\
 & Qwen3-FA & 10 & - & 0.21 $\pm$ 0.02 \\
\midrule
\multirow{5}{*}{XLSR} & Baseline & 85 & 0.09 $\pm$ 0.02 & 0.02 $\pm$ 0.01 \\
 & MFA & 82 & 0.23 $\pm$ 0.09 & 0.13 $\pm$ 0.07 \\
 & MMS-300m-IPA* & 85 & 0.23 $\pm$ 0.03 & 0.17 $\pm$ 0.02 \\
 & Wav2Vec2-IPA* & 85 & 0.22 $\pm$ 0.02 & 0.17 $\pm$ 0.02 \\
 & Qwen3-FA & 10 & - & 0.18 $\pm$ 0.01 \\
\bottomrule
\end{tabular}
}
\caption{Multilingual evaluation of forced alignment using PCMI and WACS across 85 languages from FLEURS dataset. \textbf{Emb.} denotes representation model used. *Trained in this work.}
\label{tab:headline_metrics}
\end{table}

To address this gap, we propose two reference-free corpus-level evaluation metrics based on representations from self-supervised (SSL) speech models: \emph{Phoneme-Cluster Mutual Information (PCMI)} and \emph{Word Acoustic Consistency Score (WACS)}. PCMI measures consistency between aligned phoneme labels and emergent clusters in SSL-speech representations, while WACS measures acoustic consistency of word occurrences using dynamic time warping \citep{sakoe1978dynamic} over representation sequences. The proposed metrics do not require gold timestamp annotations.

We analyze the behavior of the metrics under random perturbations of Buckeye alignments \citep{pitt2005buckeye}. Additionally, we investigate systematic adversarial perturbations such as vowel and silence absorption to characterize the scope and limitations of the metrics. We find that both PCMI and WACS degrade consistently under increasing perturbation, while WACS remains relatively insensitive to silence absorption due to the properties of dynamic time warping.

To study multilingual behavior, we train multilingual IPA-based phoneme recognition models on 85 FLEURS languages \citep{conneau2023fleurs} and generate alignments using connectionist temporal classifier (CTC) segmentation \citep{kurzinger2020ctc}. Comparing these alignments against MFA and evenly spaced phoneme baselines, the proposed metrics reveal a clear bimodal distribution for MFA alignments across languages, separating failed and successful alignments, while CTC-based aligners exhibit greater stability (Table~\ref{tab:headline_metrics}). We additionally validate the metrics on 45 languages from the DoReCo corpus \citep{paschen-etal-2020-building, doreco-2-0}, demonstrating strong agreement with manually annotated word-level and phone-level alignments.

Finally, we validate the proposed metrics against manually annotated alignments for two low-resource as well as phonologically complex languages Archi and Rutul, using recent phoneme models and data for these languages \citep{akavarapu-etal-2026-hard}. These experiments further validate the robustness of PCMI and clarify the operational limitations of WACS under silence-related perturbations.

Our contributions are threefold:
\begin{enumerate}
    \item We propose PCMI and WACS, reference-free metrics for forced alignment evaluation based on SSL-speech representations.
    \item We study the robustness and adversarial behavior of the metrics under both random and systematic perturbations.
     \item We conduct large-scale multilingual analysis across 85 FLEURS languages, validate against manually annotated alignments from 45 DoReCo languages and two phonologically complex low-resource languages. We release multilingual phoneme recognition models used for alignment as a byproduct.\footnote{\url{https://hf.co/mahesh27/mms-300m-ipa-fleurs}\\ \url{https://hf.co/mahesh27/mms-300m-xsampa-doreco}}
\end{enumerate}

\section{Related Work}
Forced alignment quality is traditionally evaluated against manually annotated timestamps using boundary displacement or overlap-based measures. Prior work commonly reported the proportion of boundaries within fixed tolerances such as 10--50\,ms from manual annotations \citep{kelley2024mason, rousso2024tradition}, with standard thresholds for accurate alignment ranging from 20\,ms \citep{hosom2009speaker} upto 200\,ms \citep{bain2023whisperx, rastorgueva2023nemo}. Other work evaluates overlap between manually annotated and automatically aligned intervals using measures such as overlap rate \citep{gonzalez2018recursive} or average temporal shift between aligned intervals \citep{shi2022achieving}. These approaches require gold timestamp annotations and are difficult to scale to multilingual settings.

Information-theoretic measures such as mutual information and cluster purity have previously been used in speech processing for phoneme-feature selection \citep{omar2002evaluation}, speaker clustering \citep{cohen2021speaker}, and speaker recognition evaluation using normalized mutual information \citep{mridha2021u}. Pointwise mutual information between aligned strings of sound classes has also been employed to derive similarity scores for phonetic sequence alignment and phylogenetic inference \citep{jager2013phylogenetic, jager2015support}. Dynamic time warping based similarity measures are also widely used in spoken term detection and retrieval \citep{sakoe1978dynamic, tejedor2012comparison}. Motivated by these approaches, we employ mutual information and dynamic time warping to define reference-free metrics for forced alignment evaluation.

\section{Problem Formulation}
\label{sec:problem}
We consider a transcript sequence $x_1, x_2, \dots, x_L$ where each \(x_l\) corresponds to a word or phoneme token. Silence tokens (contiguous) may optionally be inserted between adjacent units as well as at the beginning and end of the utterance. An alignment for an utterance of duration \(T\) performed by an alignment model $\mathcal{A}$ is defined by a sequence of boundary times $0 = t_0 \leq t_1 \leq \dots \leq t_L = T$ where token \(x_l\) occupies the interval \([t_{l-1}, t_l]\). Given gold boundary times $0 = t_0' \leq t_1' \leq \dots \leq t_L' = T$
alignment quality is commonly measured using Average Accumulated Shift (AAS) i.e., temporal deviation between postulated and gold boundaries \citep{shi2022achieving, shi2026qwen3}:
\[ \mathrm{AAS}=\frac{1}{L}\sum_{l=1}^{L}|t_l - t_l'|.\]
Our goal is to introduce reference-free evaluation metrics whose behavior correlates with alignment quality in a manner similar to AAS, without requiring gold timestamp annotations. AAS may be computed with respect to either word-level or phoneme-level boundaries. In practice, these values differ by less than 5\,ms across the datasets considered in this work. Larger discrepancies arise primarily under artificial perturbations that modify one level of annotation while leaving the other unchanged. Unless otherwise stated, we therefore report AAS as the maximum of the word-level and phoneme-level AAS values.

\section{PCMI and WACS}
\label{sec:metrics}

\subsection{Phoneme-Cluster Mutual Information (PCMI)}
\label{subsec:pcmi}

SSL-speech representations learned through predictive and clustering-based pretraining objectives \citep{baevski2020wav2vec, hsu2021hubert} are known to encode phonetic structure, including articulation and phonological attributes \citep{cormac-english-etal-2022-domain, de2024layer}. We exploit this property to evaluate forced alignments by measuring consistency between aligned phoneme labels and clusters induced from these representations.

Consider utterances $S_1,\ldots,S_N$, where each utterance $S_n$ is associated with aligned phoneme tokens $p_1,\ldots,p_{L_n}$ including silences, together with predicted boundary times from an aligner $\mathcal{A}$ (\S~\ref{sec:problem}). Given a self-supervised model $\mathcal{M}$, we extract frame-level representations $\mathbf{emb}_1,\ldots,\mathbf{emb}_F$ from a fixed layer over a sample of frames of size $F$ and assign each frame the phoneme label corresponding to the aligned interval containing the frame timestamp. We then cluster the representations using $K$-Means to obtain cluster assignments $c_1,\ldots,c_F$. PCMI is defined as the normalized mutual information between phoneme labels $P$ and representation clusters $C$:
\[
\mathrm{PCMI}=\frac{I(P;C)}{\sqrt{H(P)H(C)}},
\]
where $H$ is entropy and $I$ denotes mutual information given by:
\[I(P;C) = H(P) - H(P \mid C)\]
High PCMI indicates that the alignment induces phoneme assignments consistent with emergent phonetic structure in the representation space. Under sufficiently discriminative representations and approximately stationary phoneme realizations, increasing alignment boundary error introduces phoneme-label corruption near segment boundaries, which monotonically increases conditional entropy $H(P \mid C)$ and therefore decreases PCMI.

\subsection{Word Acoustic Consistency Score (WACS)}
\label{subsec:wacs}
SSL-speech models are known to encode substantial word-level information and support acoustic word discrimination tasks \citep{pasad-etal-2024-self, meghanani-hain-2024-improving}. Such representations have also been used for speech alignment through dynamic time warping \citep{zhu-etal-2024-taste, sakoe1978dynamic}. Motivated by these observations, WACS measures whether repeated occurrences of the same aligned word exhibit greater acoustic consistency than unrelated word pairs.

For each aligned word occurrence, we extract the sequence of frame-level representations spanning its aligned interval,
$\mathbf{e}_w=\{\mathbf{emb}_1,\ldots,\mathbf{emb}_{|w|}\}$.
Given two word occurrences $w_1$ and $w_2$, we compute a similarity score
$\mathrm{dtw}(\mathbf{e}_{w_1},\mathbf{e}_{w_2})$ using dynamic time warping (DTW)
with cosine similarity between frame-level representations (Appendix~\ref{app:wacs_details}).

Let $\mathcal{F}$ denote the set of all unique word forms in the corpus, and let
$\Sigma^+_f$ denote the set of all word occurrences having word form $f \in \mathcal{F}$. Similarly, let $\Sigma^-_f$ denote the set of word occurrences that have non-identical word form as $f$ (negative samples, $\Sigma^+_f \cap \Sigma^-_f = \varnothing$). WACS is defined as:
\[\mathrm{WACS}
=
\mathbb{E}_{f\sim\mathcal{F},\, (w_1, w_2)\sim \Sigma^+_f \times \Sigma^+_f}
\!\left[\mathrm{dtw}(\mathbf{e}_{w_1},\mathbf{e}_{w_2})\right] \]
\[-
\mathbb{E}_{f\sim\mathcal{F},\, (w_1, w_2)\sim \Sigma^+_f \times \Sigma^-_f}
\!\left[\mathrm{dtw}(\mathbf{e}_{w_1},\mathbf{e}_{w_2})\right]\]
Positive pairs are sampled from distinct occurrences of the same word form. High WACS indicates strong acoustic consistency within identical word forms compared to non-identical pairings.
\section{Experiments}
\label{sec:exp}

We evaluate the proposed metrics using SSL-speech representations from MMS \citep{pratap2024scaling} and XLSR \citep{conneau2021unsupervised}. We additionally compare against conventional MFCCs\footnote{Computed using a 25 ms window, 10 ms hop size, and 20 coefficients.} in the perturbation experiments as a classical acoustic baseline. We use representations from layer 15 selected based on the perturbation analysis described below. For computational efficiency, PCMI and WACS are computed on random subsets of utterances, where each subset contains 50 utterances in the case of PCMI and 200 utterances in the case of WACS. Reported perturbation results are averaged over 5 independent samples. Additional details are provided in Appendix~\ref{app:impl}.

\subsection{Perturbation Analysis}
\label{subsec:perturb}

\begin{figure*}[t]
    \centering

    \begin{minipage}{0.36\textwidth}
        \centering
        \includegraphics[width=\linewidth]{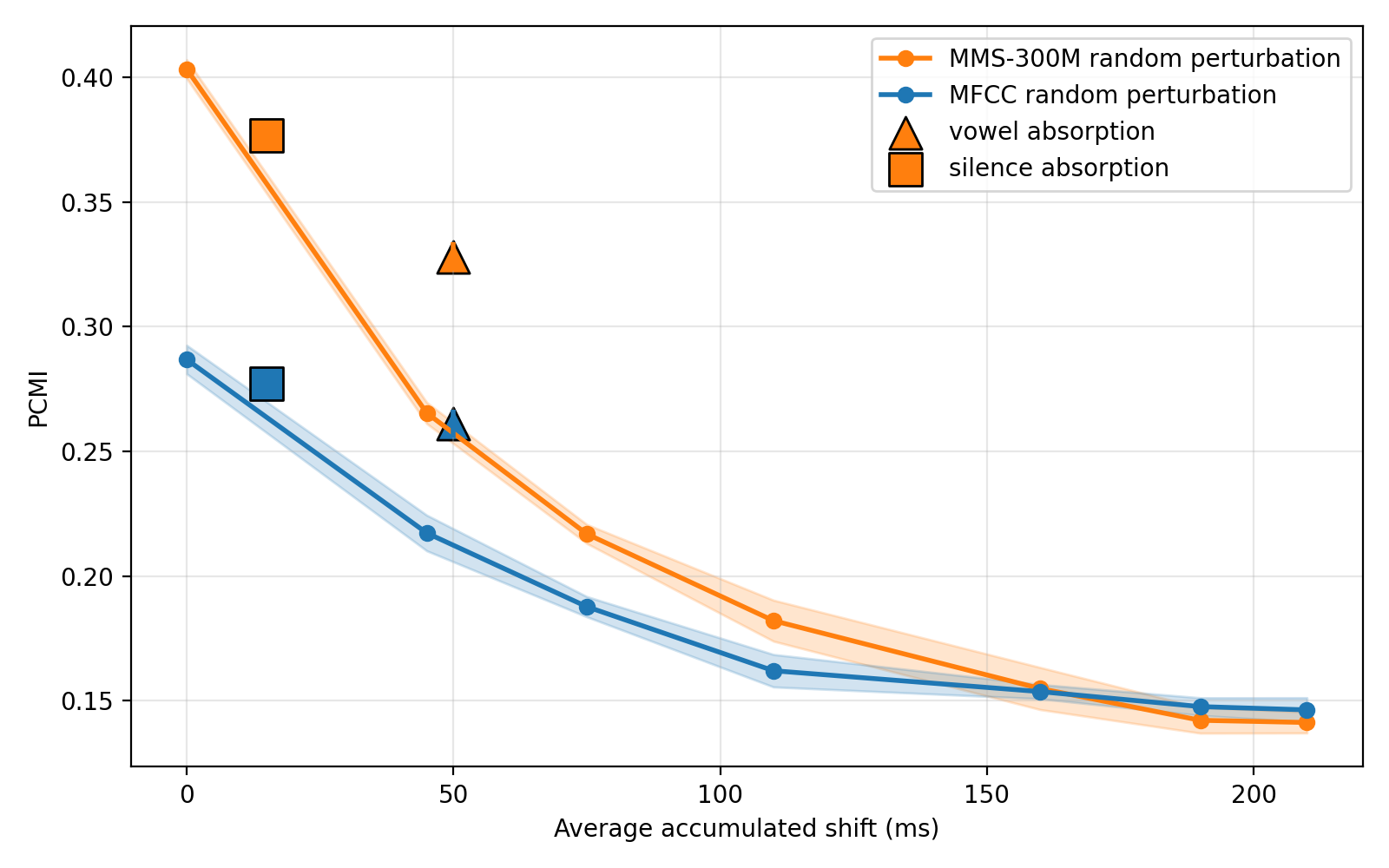}
    \end{minipage}
    \hspace*{1cm}
    \begin{minipage}{0.36\textwidth}
        \centering
        \includegraphics[width=\linewidth]{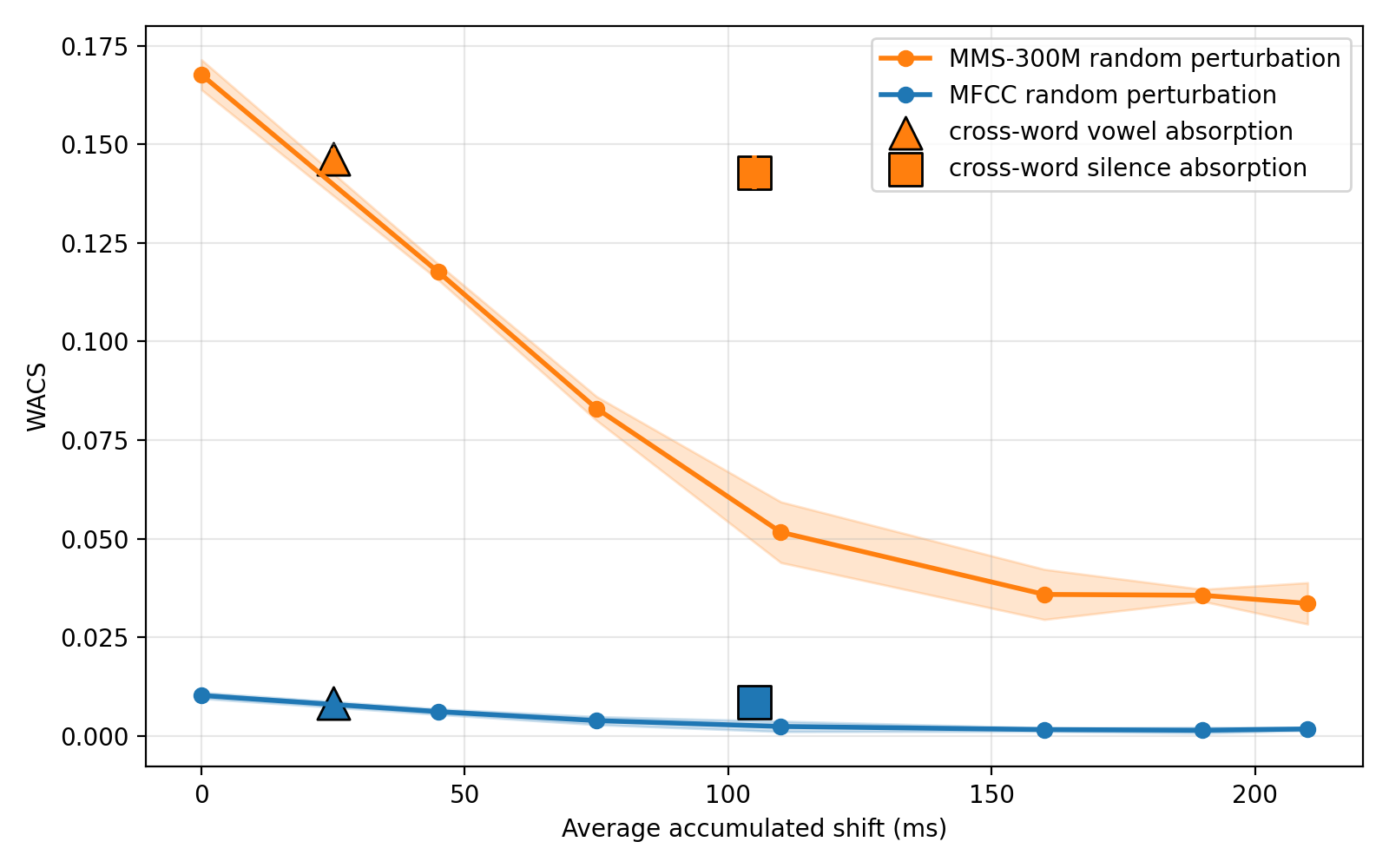}
    \end{minipage}

    \caption{
    Behavior of PCMI and WACS under random and systematic perturbations of Buckeye alignments using MMS-300M and MFCC representations. Curves show mean scores over 5 random perturbation samples, with shaded regions denoting standard deviation.
    }
    \label{fig:perturb}
\end{figure*}

We study the behavior of PCMI and WACS under controlled perturbations of manually annotated alignments from the Buckeye corpus \citep{pitt2005buckeye}, which contains word- and phoneme-level annotations for 40 speakers. We segment the recordings into utterances of duration 3--20\,s, yielding 2935 utterances totaling about 7.5 hours.

\paragraph{Random perturbations.}
To simulate progressively degraded alignments, we perturb word boundaries by adding Gaussian noise with target standard deviation $\sigma$. Boundary constraints are then reimposed to avoid overlaps and intervals with zero-duration, after which phoneme boundaries are warped proportionally within each word interval. Although perturbations were generated with target scales ranging from 50--2000\,ms, the resulting average accumulated shift (AAS) after constraint enforcement ranged from approximately 45--210\,ms. Figure~\ref{fig:perturb} reports the resulting behavior of PCMI and WACS using MMS-300M and MFCC representations.

Both PCMI and WACS degrade consistently with increasing perturbation severity. The metrics decrease approximately linearly at lower perturbation levels before gradually plateauing at larger shifts. While the overall degradation trends remain similar across MMS-300M and MFCC representations, notable differences emerge in score ranges. PCMI retains a comparatively broad operating range under MFCC features, suggesting that local phonetic structure remains sufficiently recoverable from conventional spectral representations. In contrast, WACS values under MFCC representations occupy a substantially narrower range, whereas self-supervised MMS representations produce considerably stronger separations. This suggests that WACS particularly benefits from the richer lexical and contextual structure captured by SSL-speech representations.

\begin{figure*}[t]
    \centering
    \includegraphics[width=0.48\linewidth]{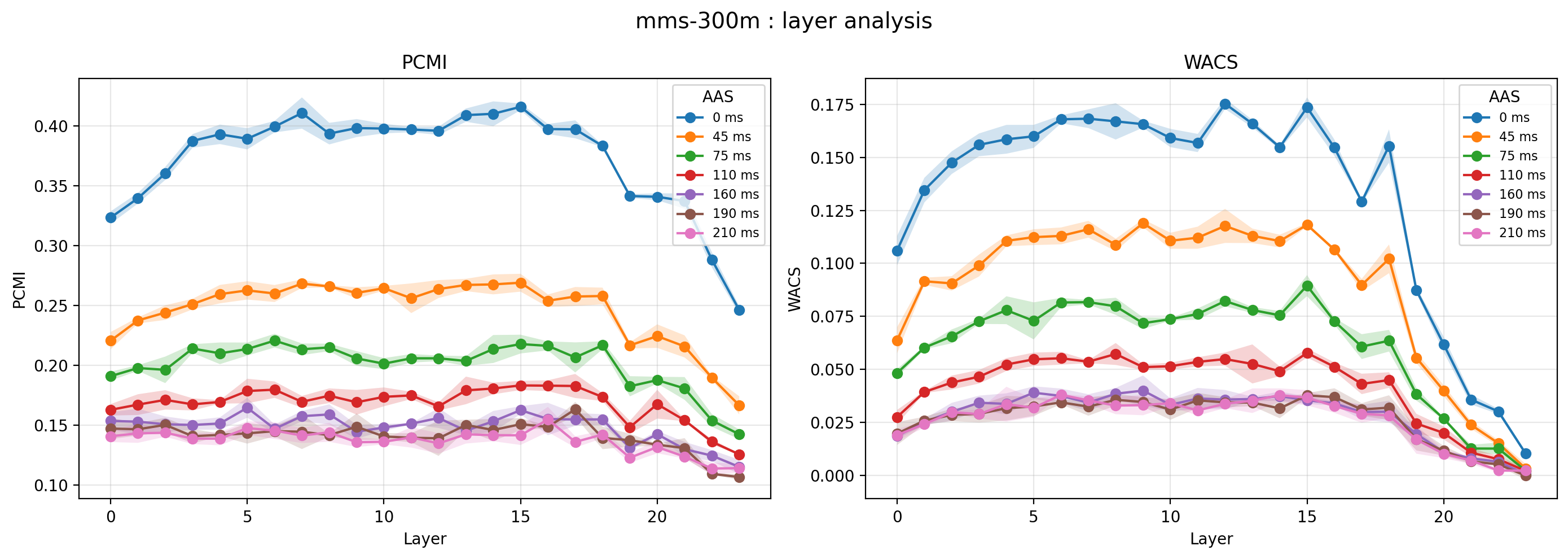}~
    \includegraphics[width=0.48\linewidth]{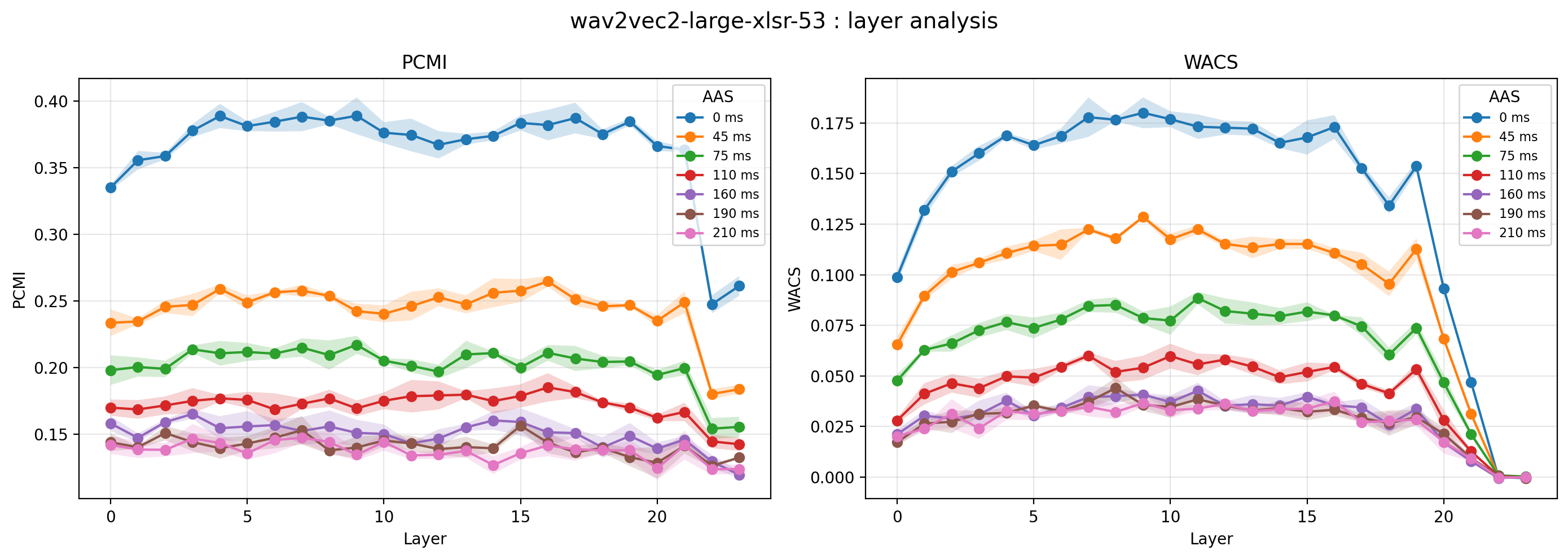}
    \caption{Layer-wise behavior of PCMI and WACS under increasing perturbation severity using MMS (left) and XLSR (right) representations. Middle transformer layers provide the strongest separation between clean and perturbed alignments, while later layers exhibit reduced sensitivity.}
    \label{fig:layers}
\end{figure*}

We additionally analyze sensitivity across network layers using representations extracted from hidden states of different transformer blocks of MMS and XLSR. As shown in Figure~\ref{fig:layers}, middle layers exhibit the clearest separation between low- and high-AAS alignments for both PCMI and WACS, whereas early layers may be dominated by local acoustic variation and later layers may become increasingly contextualized. This behavior is consistent with prior analyses showing that intermediate layers of SSL-speech models encode enhanced phonetic structure \citep{cormac-english-etal-2022-domain, de2024layer}.

Across both metrics, layers approximately between 5 and 16 show the greatest perturbation sensitivity. We therefore select layer 15 for all subsequent experiments by maximizing the separation between clean alignments and the most severely perturbed alignments.

We further analyze the sensitivity of PCMI to the number of K-means clusters used for representation quantization. The separation between clean and severely perturbed alignments remains stable, varying only from approximately $0.26$ to $0.25$ for MMS and from $0.24$ to $0.22$ for XLSR representations across $25 \leq \mathrm{n\_cluster} \leq 100$ (see Appendix Figure~\ref{fig:clusters}). We therefore select $\mathrm{n\_cluster}=50$ for all subsequent experiments, which approximately matches the average phoneme inventory (labels) (Appendix Table~\ref{tab:multilingual-stats}).

\paragraph{Systematic perturbations.}
Random perturbations alone do not capture structured alignment errors that may interact favorably with representation-based metrics. We therefore construct adversarial perturbations motivated by phonological regularities and DTW behavior.

First, we introduce vowel absorption perturbations by merging vowel-consonant transitions, motivated by the tendency of neighboring phonetic segments to exhibit similar acoustic structure. These perturbations primarily affect PCMI by altering phoneme assignments while preserving coarse acoustic continuity.

Second, we construct silence absorption perturbations by merging preceding silence intervals into neighboring words. Since DTW is relatively insensitive to leading or trailing silence, this perturbation particularly targets WACS.

The resulting perturbations achieve moderate AAS while producing substantially weaker degradation than random perturbations at comparable shift levels. In particular, silence absorption emerges as a prominent adversarial failure mode for WACS. 

Nevertheless, both metrics remain informative under realistic alignment deviations: PCMI continues to reflect phonetic consistency under moderate perturbations, while WACS remains effective when silence regions are excluded during downstream segment extraction. Moreover, further degenerate merging of phoneme labels cannot artificially inflate PCMI, since $\mathrm{PCMI}\to0$ as the phoneme-label entropy $H(P)\to0$ (see Appendix \ref{app:pcmi-collapse}). Likewise, if the representation model collapses all word occurrences to identical embeddings, positive and negative DTW similarities become indistinguishable (see \S{\ref{subsec:wacs}}), causing $\mathrm{WACS} \rightarrow 0$.

\section{Multilingual Evaluation}
\label{sec:multieval}

We analyze PCMI and WACS in a large-scale multilingual setting, which constitutes the primary target application of the proposed metrics. The multilingual experiments reported in this section use layer-15 representations from MMS and XLSR.

\subsection{Multilingual Alignment Models}
\label{subsec:fleurs}
\begin{table}[t]
\centering
\scriptsize
\setlength{\tabcolsep}{4pt}

\begin{tabular}{lc|lclc}
\toprule
\multicolumn{2}{c|}{\textbf{FLEURS}} &
\multicolumn{4}{c}{\textbf{DoReCo}}\\
\midrule
\textbf{Family} & \textbf{\#} &
\textbf{Family} & \textbf{\#} &
\textbf{Family} & \textbf{\#} \\
\midrule
Indo-European & 41 &
Austronesian & 7 &
Mayan & 1 \\
Niger-Congo & 9 &
Sino-Tibetan & 4 &
Mixe-Zoque & 1 \\
Afroasiatic & 7 &
Indo-European & 3 &
Nilo-Saharan & 1 \\
Austronesian & 5 &
Niger-Congo & 3 &
Pama-Nyungan & 1 \\
Turkic & 5 &
Afroasiatic & 2 &
Pano-Tacanan & 1 \\
Dravidian & 4 &
Arawakan & 2 &
Trans-New Guinea & 1 \\
Sino-Tibetan & 3 &
Austroasiatic & 2 &
Tungusic & 1 \\
Uralic & 3 &
Nakh-Daghestanian & 2 &
Tuu & 1 \\
Austroasiatic & 2 &
Turkic & 2 &
Uralic & 1 \\
Kra-Dai & 2 &
Algic & 1 &
Yam & 1 \\
Japonic & 1 &
Boran & 1 &
Mixed Language & 1 \\
Kartvelian & 1 &
Chibchan & 1 &
Isolate & 1 \\
Koreanic & 1 &
Gunwinyguan & 1 \\
Mongolic & 1 &
Kartvelian & 1 \\
& &
Koreanic & 1 \\
\bottomrule
\end{tabular}

\caption{Language family distributions in the multilingual evaluation datasets --- FLEURS (85 languages) and DoReCo (45 languages).}
\label{tab:family_stats}
\end{table}
\begin{figure*}[t]
    \centering
    \includegraphics[width=0.9\linewidth]{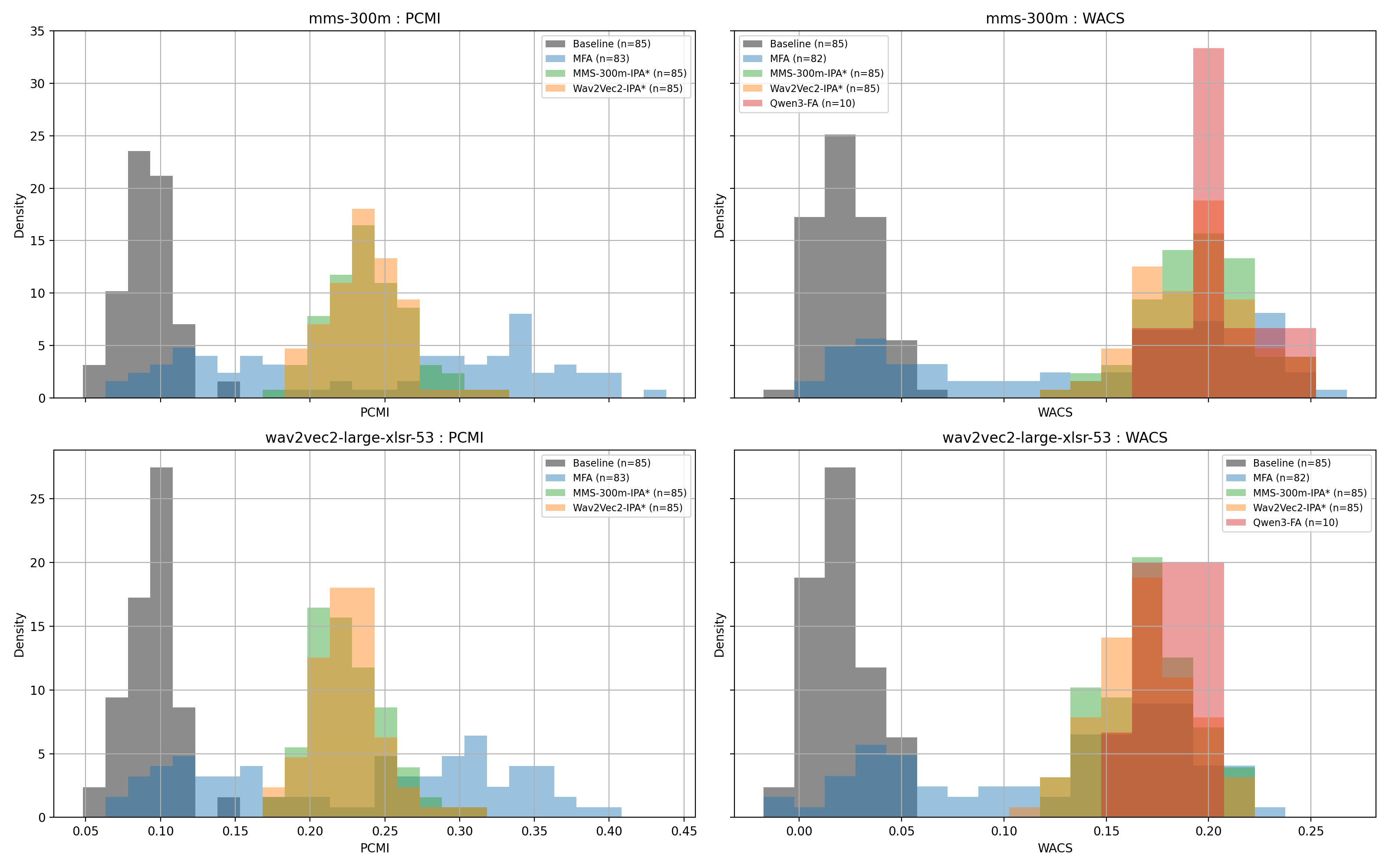}
    \caption{
    Distribution of PCMI (left) and WACS (right) across multilingual alignments on FLEURS. The top row uses MMS representations while the bottom row uses XLSR representations. MFA exhibits a clear bimodal structure corresponding to successful and failed alignments, while CTC-based aligners produce substantially more stable distributions across languages. * Trained in this work.
    }
    \label{fig:multilingual}
\end{figure*}

\paragraph{Dataset}
To analyze the proposed metrics in a multilingual setting, we use the FLEURS dataset \citep{conneau2023fleurs}, which provides speech and transcriptions spanning a diverse set of languages and writing systems. Since the phoneme-based evaluation framework requires a unified cross-lingual representation, all transcripts were converted into International Phonetic Alphabet (IPA) format. To complement this large-scale evaluation, we additionally use the DoReCo corpus \citep{doreco-2-0}, a collection of predominantly low-resource and endangered languages with manually annotated word- and phoneme-level timestamps. Unlike FLEURS, DoReCo provides gold alignment annotations, enabling direct comparison against timestamp-based evaluation measures such as AAS. The corpus includes phoneme-level transcriptions in X-SAMPA format, which are retained without further conversion throughout our experiments.

A particular challenge in FLEURS is the presence of numerals, and mixed symbolic forms within transcripts. These are often unsuitable for direct grapheme-to-phoneme conversion because their pronunciation depends on linguistic context and language-specific conventions. For example, the token ``2019'' may be realized as ``twenty nineteen'', ``two thousand nineteen''. We therefore normalize numerals into fully written forms prior to phonemization using the \texttt{num2words} package\footnote{\url{https://github.com/savoirfairelinux/num2words}}, which we further extended to support several languages missing from the original implementation. We additionally apply a simple heuristic where values between 1200 and 2050 are interpreted as years, while all remaining are treated as cardinal numbers. Grapheme-to-phoneme conversion for FLEURS data was subsequently performed using Epitran \citep{mortensen-etal-2018-epitran} and XPF \citep{XPF2021manual}, producing IPA transcriptions. 

The language set spans a broad range of families, summarized in Table~\ref{tab:family_stats}, while detailed language-wise statistics and metadata are provided in Table~\ref{tab:language_info} (FLEURS) \& Table~\ref{tab:doreco-languages} (DoReCo).

\paragraph{Phoneme Models}
Using these transcriptions, we finetune multilingual phoneme recognition models MMS-300M-IPA and Wav2Vec2-IPA, based on MMS-300M \citep{pratap2024scaling} and Wav2Vec2-Phoneme \citep{xu2022simple} respectively, for FLEURS, and MMS-300M-DORECO for DoReCo, based on MMS-300M. All models use language-specific adapters. Additional training details are provided in Appendix~\ref{app:ipa-train}. Alignments are generated using CTC segmentation \citep{kurzinger2020ctc}.

We compare the resulting alignments on FLEURS against several baselines and alignment systems:
(i) MFA \citep{mcauliffe2017montreal} acoustic models trained independently for each language,
(ii) an evenly spaced phoneme baseline obtained by uniformly partitioning non-silent regions (with only utterance-initial and utterance-final silences removed),
and (iii) Qwen3-ForcedAligner-0.6B (Qwen3-FA) \citep{shi2026qwen3}, evaluated on the subset of 10 supported languages for which word-level alignments are available.

For evaluation on DoReCo, in addition to the evenly spaced baseline, we include a silence-absorbing post-processed variant of MMS-300M-DORECO (denoted by the suffix `-SIL'). This variant is included because gold alignments are available for DoReCo, allowing AAS to be reported, and AAS is particularly sensitive to silence placement. We do not evaluate MFA on DoReCo, as the corpus provides only approximately two hours of speech per language on average, making language-specific MFA training impractical.

\paragraph{Results}
The multilingual distributions of PCMI and WACS on FLEURS are shown in Figure~\ref{fig:multilingual}. The MFA exhibits pronounced bimodal behavior in both metrics; manual inspection reveals that the lower-scoring modes predominantly correspond to failed alignments, while the higher-scoring modes correspond to successful ones. In contrast, CTC-based aligners produce substantially more stable distributions across languages despite achieving slightly lower peak PCMI values than the successful MFA cluster.

Failure rates further support this observation. MFA alignment training or decoding failed for approximately 12k out of 65k utterances ($\sim$19\%), whereas the proposed IPA-based CTC aligners produced only 11 word-level failures overall. The evenly spaced phoneme baseline remains competitive, indicating that coarse temporal consistency alone can yield moderate acoustic agreement. The distributions of metrics on the DoReCo corpus, which additionally provides manually annotated word- and phone-level boundaries, are provided in Appendix Figure~\ref{fig:multilingual_doreco}. Manually annotated gold alignments achieve average scores of approximately 0.33 PCMI and 0.13 WACS, providing practical reference values for high-quality alignments.

The metrics exhibit similar behavior across MMS and XLSR representations, with MMS providing slightly stronger separation overall. Notably, XLSR remains effective despite being pretrained on only 53 languages---with many of the evaluated FLEURS languages and most of DoReCo absent from its training data---whereas MMS pretraining substantially overlaps with the FLEURS language inventory, yet rarely with DoReCo. This suggests that the proposed metrics are reasonably robust across different SSL representation models.

Aggregated multilingual statistics, including cluster purity, vocabulary sizes, pair counts, entropy measures, and score variances, are summarized in Table~\ref{tab:multilingual-stats}. Across languages and embedding models, normalized cluster entropy consistently remains high (approximately $0.95$--$0.99$), indicating effective utilization of the cluster inventory by the representation space. Normalized phoneme-label entropy also remains relatively high (roughly $0.82$--$0.95$), reflecting substantial phonetic diversity across the evaluated languages. MFA exhibits substantially larger variance in these statistics as well as in PCMI and WACS scores, consistent with its observed bimodal failure behavior. In contrast, CTC-based aligners show considerably tighter distributions.

Language-wise alignment metrics are reported in Appendix Tables \ref{tab:pcmi_table}, \ref{tab:wacs_table}, \ref{tab:pcmi_table_doreco}, \ref{tab:wacs_table_doreco} and \ref{tab:aas_table_doreco}. Language-wise phoneme recognition performance in terms of word and character error rates (WER, CER) is reported in Appendix Tables \ref{tab:multilingual_asr} and \ref{tab:doreco_asr}.

\begin{table}[t]
\centering
\resizebox{0.9\columnwidth}{!}{
\begin{tabular}{llcc}
\toprule
\textbf{Embedding} & \textbf{Metric Pair} & \textbf{Pearson's ~\textit{r}} & \textbf{p-value} \\
\midrule
\multirow{2}{*}{MMS} & PCMI vs AAS & -0.7769 & $<0.001^{***}$ \\
& WACS vs AAS & -0.6257 & $<0.001^{***}$ \\
\midrule
\multirow{2}{*}{XLSR} & PCMI vs AAS & -0.7761 & $<0.001^{***}$ \\
& WACS vs AAS & -0.6686 & $<0.001^{***}$ \\
\bottomrule
\end{tabular}
}
\caption{Pearson correlation between reference-free metrics (PCMI, WACS) and alignment quality measured by AAS on the 45-language DoReCo evaluation set. Lower AAS indicates better alignment quality.}
\label{tab:doreco_corr_aas}
\end{table}
\paragraph{Correlations with AAS}
The availability of manually annotated word- and phone-level boundaries in DoReCo additionally allows direct comparison of the proposed reference-free metrics against Average Accumulated Shift (AAS), a timestamp-based measure of alignment quality. Table~\ref{tab:doreco_corr_aas} reports Pearson correlations between PCMI/WACS and AAS across 45 languages and four alignment conditions (gold, baseline, MMS-300M-DORECO, and MMS-300M-DORECO-SIL), yielding 180 alignment sets in total.

Both PCMI and WACS exhibit strong negative correlations with AAS across MMS and XLSR representations, indicating that higher metric scores consistently correspond to lower alignment error. PCMI shows the strongest relationship overall ($r\sim-0.78$), while WACS provides complementary evidence with correlations between $r=-0.62$ and $-0.67$. These results provide large-scale multilingual validation against manually annotated timestamps and further support the effectiveness of the proposed metrics as reference-free indicators of alignment quality.

\begin{table*}[t]
\centering
\scriptsize
\begin{minipage}{0.49\linewidth}
\centering
\textbf{WER Correlations}\\
\vspace{2mm}

\resizebox{\textwidth}{!}{
\begin{tabular}{llcc}
\toprule
\textbf{Aligner} & \textbf{Embedding} & \textbf{PCMI} & \textbf{WACS} \\
\midrule

\multirow{2}{*}{MMS-300M-IPA*}
& MMS & -0.161 / -0.137 & -0.294 / -0.308 \\
& XLSR & -0.092 / -0.050 & -0.355 / -0.363 \\
\midrule

\multirow{2}{*}{Wav2Vec2-IPA*}
& MMS & -0.008 / -0.073 & -0.027 / -0.111 \\
& XLSR  & -0.001 / -0.108 & -0.152 / -0.212 \\
\midrule

\multirow{2}{*}{MMS-300M-DORECO*}
& MMS & -0.200 / -0.225 & -0.259 / -0.237 \\
& XLSR & -0.225 / -0.214 & -0.261 / -0.230 \\
\bottomrule
\end{tabular}}
\end{minipage}
\hfill
\begin{minipage}{0.49\linewidth}
\centering
\textbf{CER Correlations}\\
\vspace{2mm}

\resizebox{\textwidth}{!}{
\begin{tabular}{llcc}
\toprule
\textbf{Aligner} & \textbf{Embedding} & \textbf{PCMI} & \textbf{WACS} \\
\midrule

\multirow{2}{*}{MMS-300M-IPA*}
& MMS & -0.351 / -0.334 & -0.529 / -0.533 \\
& XLSR & -0.263 / -0.226 & -0.538 / -0.552 \\
\midrule

\multirow{2}{*}{Wav2Vec2-IPA*}
& MMS & -0.398 / -0.383 & -0.438 / -0.441 \\
& XLSR  & -0.385 / -0.413 & -0.475 / -0.481 \\
\midrule

\multirow{2}{*}{MMS-300M-DORECO*}
& MMS & -0.562 / -0.616 & -0.459 / -0.318 \\
& XLSR & -0.547 / -0.552 & -0.483 / -0.316 \\
\bottomrule
\end{tabular}}
\end{minipage}

\caption{Correlation between ASR error rates (WER, CER) and alignment metrics (PCMI, WACS). Values are Pearson/Spearman correlations. * trained in this work}
\label{tab:corr_asr_side_by_side}
\end{table*}
\paragraph{Phoneme recognition and alignment quality}
We further analyze the relationship between phoneme recognition performance and alignment quality across languages. As shown in Table~\ref{tab:corr_asr_side_by_side}, character error rate (CER) exhibits consistent moderate negative correlations with alignment metrics, with correlation magnitudes of approximately $-0.5$ with WACS and $-0.3\sim-0.5$ with PCMI across embedding and alignment models. This indicates that improvements in phoneme recognition quality are generally associated with improved alignment quality, although the relationship is not strict.

WER shows a similar but weaker trend, with moderate negative correlation with WACS for MMS-300M-IPA ($\sim-0.3$), while remaining weak for several configurations. This suggests that character-level evaluation is a more reliable predictor of alignment quality than word-level error rates in this setting.

As reported in Appendix Tables \ref{tab:multilingual_asr} and \ref{tab:doreco_asr}, WER values are generally high across languages, particularly for Wav2Vec2-IPA models and for DoReCo corpus, partly reflecting the absence of explicit language modeling. Despite this, alignment quality remains relatively robust, indicating that forced alignment can greatly compensate for phoneme recognition noise.

Overall, these results suggest a consistent but moderate coupling between phoneme recognition accuracy and alignment quality, with stronger and more stable effects observed for CER compared to WER. Together with the strong negative correlations between PCMI/WACS and AAS on DoReCo (Table~\ref{tab:doreco_corr_aas}), these findings support the validity of the proposed metrics as indicators of alignment quality rather than mere proxies for phoneme recognition performance.

\subsection{Two phonologically complex languages}
\label{subsec:northcaucasus}

\begin{table*}[t]
\centering
\resizebox{0.6\linewidth}{!}{%
\begin{tabular}{llrcccc}
\toprule
\multirow{2}{*}{\textbf{Language}} &
\multirow{2}{*}{\textbf{Aligner}} &
\multirow{2}{*}{\textbf{AAS}$\downarrow$} &
\multicolumn{2}{c}{\textbf{MMS}} &
\multicolumn{2}{c}{\textbf{XLSR}} \\
\cmidrule(lr){4-5}
\cmidrule(lr){6-7}
& & &
\textbf{PCMI}$\uparrow$ &
\textbf{WACS}$\uparrow$ &
\textbf{PCMI}$\uparrow$ &
\textbf{WACS}$\uparrow$ \\
\midrule

\multirow{4}{*}{\textbf{Archi}}
& Gold            &   0.0  & --    & 0.225 & --    & 0.220 \\
& CTC-IPA-Archi        &  88.5  & 0.306 & 0.203 & 0.307 & 0.194 \\
& CTC-IPA-Archi - SIL  &  46.3  & 0.319 & 0.215 & 0.330 & 0.215 \\
& Baseline        &  85.8  & 0.204 & 0.112 & 0.203 & 0.102 \\

\midrule

\multirow{4}{*}{\textbf{Rutul}}
& Gold            &   0.0  & --    & 0.182 & --    & 0.182 \\
& CTC-IPA-Rutul        &  68.0  & 0.292 & 0.172 & 0.278 & 0.173 \\
& CTC-IPA-Rutul - SIL  &  61.9  & 0.292 & 0.175 & 0.277 & 0.172 \\
& Baseline        & 180.6  & 0.126 & 0.041 & 0.127 & 0.040 \\

\bottomrule
\end{tabular}
}
\caption{Evaluation against manually annotated alignments for Archi and Rutul. Lower AAS is better, while higher PCMI and WACS are better.}
\label{tab:archi-rutul}
\end{table*}

The proposed metrics are further evaluated on two endangered East Caucasian languages: Archi \citep{kibrik2007Archi} and Kina Rutul \citep{alekseevaetal2024}, using manually annotated word-level alignments produced by a linguistically trained annotator, the second author of this work. Speech data and language-specific IPA phoneme recognition models are obtained from \citet{akavarapu-etal-2026-hard}. These CTC-based IPA models were trained separately for Archi and Rutul by the original authors and are used directly in our experiments. Each language has only approximately 45--75 minutes of available training audio in the current setup, along with around 7 minutes of test data (100 test utterances in Archi, 90 in Rutul), making conventional MFA training impractical.

Results are shown in Table~\ref{tab:archi-rutul}. Despite the phonological complexity and limited supervision, the alignments after trailing-silence removal achieve AAS values roughly between 45--70 ms, which is competitive given the 20 ms frame resolution.

Manual inspection additionally revealed systematic silence absorption in several alignments. We therefore apply the silence-correction postprocessing step after CTC segmentation (denoted CTC-IPA-$\bullet$ - SIL in Table~\ref{tab:archi-rutul}). This procedure removes approximately 44 seconds of excess silence across the Archi dataset, reducing AAS from 88.5 ms to 46.3 ms, while producing only minor changes in WACS and PCMI. In contrast, only around 5 seconds of silence are removed for Rutul, leading to comparatively smaller AAS changes. This difference between Rutul and Archi is likely related to the higher speech rate in the Rutul recordings, which consist primarily of spontaneous narratives, compared to the Archi recordings, which are prepared readings of written texts.

These observations are consistent with the perturbation experiments in \S\ref{subsec:perturb}, where WACS was found to be relatively insensitive to silence-related errors compared to AAS. Consequently, while the proposed metrics generally track alignment quality well, they do not penalize silence absorption as strongly as the timestamp-based measure AAS.

Interestingly, the evenly spaced baseline performs relatively well on Archi compared to Rutul, particularly under PCMI which may be attributed to the well-articulated nature of the Archi read-out recordings. A similar phenomenon is observed for a small number of languages in DoReCo (Appendix Figure~\ref{fig:multilingual_doreco}). Importantly, this does not indicate a weakness of the proposed metrics, since these cases also exhibit correspondingly low AAS values, suggesting that the baseline alignments are genuinely competitive rather than being incorrectly favored by the metrics.

\paragraph{Systematic errors}
Beyond silence absorption, expert analysis also identified smaller but systematic alignment errors, such as the merging of the release of the consonant into the following phoneme onsets (e.g. si\textbf{k'{\super j}}$\rightarrow$wu\textipa{K}at'i, na\textbf{\textipa{\t{ts}'}}$\rightarrow$jara\textipa{K}, ja\textbf{k}$\rightarrow$uwk'uli) and of the last segment of the vowel into the onset of the following consonant (eg. uq'ul{\super j}\textbf{a}$\rightarrow$zamanindexa, minowd\textbf{\textbari}$\rightarrow$\textscg u\textipa{S}imakan). As also suggested by the perturbation results in \S\ref{subsec:perturb}, the proposed metrics are less sensitive to such fine-grained systematic mergers. These effects likely contribute to residual differences in AAS that are not explained by silence removal alone.

\section{Discussion}

Although MFA achieves high PCMI scores when alignment succeeds (Appendix~ Table~\ref{tab:pcmi_table}), it exhibits a substantial failure rate of approximately 19\%, in contrast to the near-complete robustness of CTC-based models (\S\ref{subsec:fleurs}). Moreover, MFA training is often impractical in highly under-resourced settings (\S\ref{subsec:northcaucasus}). These limitations suggest that the common practice of reporting alignment performance primarily as deviations from MFA-based alignments is not reliable in low-resource scenarios. The proposed metrics, PCMI and WACS, provide a more stable alternative for such settings.

Furthermore, AAS is highly sensitive to silence absorption effects, whereas PCMI and WACS remain comparatively robust to such perturbations (\S\ref{subsec:perturb}, \S\ref{subsec:northcaucasus}). This distinction should be viewed as a limitation of AAS rather than a disadvantage of the proposed metrics, particularly in downstream applications such as retrieval, where excessive silence can distort temporal alignment quality. In fact, manual annotations in DoReCo often do not precisely trim short silence intervals, suggesting that fine-grained silence placement is of secondary importance relative to the alignment of linguistic units themselves. Since silence can be handled deterministically through simple postprocessing or segmentation heuristics, sensitivity to silence may introduce unnecessary variance without reflecting true alignment quality. 

Similarly, AAS is also expected to be sensitive to differences in speech rate, potentially complicating direct cross-lingual comparisons. In contrast, PCMI and WACS are comparatively rate-invariant by design. Overall, the proposed metrics provide scalable and reliable alternatives for multilingual forced alignment evaluation without requiring manually annotated timestamps.

\section{Conclusions}

We introduced two reference-free corpus-level metrics for forced alignment evaluation based on SSL-speech representations: Phoneme-Cluster Mutual Information (PCMI) and Word Acoustic Consistency Score (WACS). Across both synthetic perturbations and multilingual evaluation on 85 FLEURS, 45 DoReCo and two phonologically complex languages, the metrics exhibit strong negative correlation with manual timestamp-based alignment quality measures. We further analyzed systematic effects such as silence absorption and phonological mergers, showing complementary behavior between PCMI, WACS and AAS. Unlike approaches relying on tediously manually annotated timestamps or MFA-derived references, the proposed framework enables scalable evaluation in multilingual and low-resource settings. More broadly, these results suggest that SSL-speech representations can support large-scale multilingual forced alignment evaluation and facilitate the development of alignment resources for under-resourced languages.

\newpage
\section*{Limitations}

The proposed metrics are intended as scalable surrogate measures of alignment quality rather than direct replacements for manually annotated timestamps. Nevertheless, they provide an efficient mechanism for screening model-generated alignments before manual inspection and correction. Moreover, the proposed measures operate at the corpus level and do not directly provide diagnostic information for individual utterances or local alignment errors.

Additionally, although the perturbation experiments cover both random and systematic alignment errors, certain linguistically fine-grained phenomena---such as systematic phonological mergers or other coarticulatory effects---may not be fully captured by the proposed metrics.

Finally, while we evaluate alignments across 132 languages and multiple alignment systems, the experiments are restricted to datasets with available transcriptions and phoneme conversion pipelines. Languages without reliable grapheme-to-phoneme resources remain comparatively underexplored. We additionally do not ablate the effect of different G2P systems, since for most evaluated languages no viable alternative phonemization resources are available.

\section*{Ethics Statement}

All datasets used in this work were obtained from publicly available sources and used in accordance with their respective licenses and terms of use. To the best of our knowledge and based on the available dataset documentation, the data do not contain personally identifying information or intentionally offensive content. The manually produced word-level alignment annotations likewise do not present foreseeable ethical concerns, as they consist solely of temporal boundary labels over already publicly available speech recordings. Large language models and AI-assisted coding tools, including ChatGPT and GitHub Copilot, were used to assist with portions of code development, and manuscript refinement. All generated outputs were carefully reviewed and verified by the authors. The core research ideas, experimental design, analyses, and conclusions are the authors' own, although selected refinements suggested by these tools were incorporated where appropriate.

\section*{Acknowledgments}
Mahesh Akavarapu received funding from Volkswagen Foundation under the Phylomilia project within the Pioneering Projects funding line. We thank the anonymous reviewers, whose feedback significantly improved the paper.

\bibliography{anthology,custom}

@inproceedings{mortensen-etal-2018-epitran,
    title = "{E}pitran: Precision {G}2{P} for Many Languages",
    author = "Mortensen, David R.  and
      Dalmia, Siddharth  and
      Littell, Patrick",
    editor = "Calzolari, Nicoletta  and
      Choukri, Khalid  and
      Cieri, Christopher  and
      Declerck, Thierry  and
      Goggi, Sara  and
      Hasida, Koiti  and
      Isahara, Hitoshi  and
      Maegaard, Bente  and
      Mariani, Joseph  and
      Mazo, H{\'e}l{\`e}ne  and
      Moreno, Asuncion  and
      Odijk, Jan  and
      Piperidis, Stelios  and
      Tokunaga, Takenobu",
    booktitle = "Proceedings of the Eleventh International Conference on Language Resources and Evaluation ({LREC} 2018)",
    month = may,
    year = "2018",
    address = "Miyazaki, Japan",
    publisher = "European Language Resources Association (ELRA)",
    url = "https://aclanthology.org/L18-1429/"
}

@inproceedings{cormac-english-etal-2022-domain,
    title = "Domain-Informed Probing of wav2vec 2.0 Embeddings for Phonetic Features",
    author = "Cormac English, Patrick  and
      Kelleher, John D.  and
      Carson-Berndsen, Julie",
    editor = "Nicolai, Garrett  and
      Chodroff, Eleanor",
    booktitle = "Proceedings of the 19th SIGMORPHON Workshop on Computational Research in Phonetics, Phonology, and Morphology",
    month = jul,
    year = "2022",
    address = "Seattle, Washington",
    publisher = "Association for Computational Linguistics",
    url = "https://aclanthology.org/2022.sigmorphon-1.9/",
    doi = "10.18653/v1/2022.sigmorphon-1.9",
    pages = "83--91"
}

@inproceedings{ahn-chodroff-2022-voxcommunis,
    title = "{V}ox{C}ommunis: A Corpus for Cross-linguistic Phonetic Analysis",
    author = "Ahn, Emily  and
      Chodroff, Eleanor",
    editor = "Calzolari, Nicoletta  and
      B{\'e}chet, Fr{\'e}d{\'e}ric  and
      Blache, Philippe  and
      Choukri, Khalid  and
      Cieri, Christopher  and
      Declerck, Thierry  and
      Goggi, Sara  and
      Isahara, Hitoshi  and
      Maegaard, Bente  and
      Mariani, Joseph  and
      Mazo, H{\'e}l{\`e}ne  and
      Odijk, Jan  and
      Piperidis, Stelios",
    booktitle = "Proceedings of the Thirteenth Language Resources and Evaluation Conference",
    month = jun,
    year = "2022",
    address = "Marseille, France",
    publisher = "European Language Resources Association",
    url = "https://aclanthology.org/2022.lrec-1.566/",
    pages = "5286--5294"
}

@inproceedings{akavarapu-etal-2026-hard,
    title = "Hard to Be Heard: Phoneme-Level {ASR} Analysis of Phonologically Complex, Low-Resource Endangered Languages",
    author = {Akavarapu, V.S.D.S.Mahesh  and
      Daniel, Michael  and
      J{\"a}ger, Gerhard},
    editor = "Liakata, Maria  and
      Moreira, Viviane P.  and
      Zhang, Jiajun  and
      Jurgens, David",
    booktitle = "Findings of the {A}ssociation for {C}omputational {L}inguistics: {ACL} 2026",
    month = jul,
    year = "2026",
    address = "San Diego, California, United States",
    publisher = "Association for Computational Linguistics",
    url = "https://aclanthology.org/2026.findings-acl.147/",
    doi = "10.18653/v1/2026.findings-acl.147",
    pages = "3014--3028",
    ISBN = "979-8-89176-395-1"
}

@inproceedings{paschen-etal-2020-building,
    title = "Building a Time-Aligned Cross-Linguistic Reference Corpus from Language Documentation Data ({D}o{R}e{C}o)",
    author = "Paschen, Ludger  and
      Delafontaine, Fran{\c{c}}ois  and
      Draxler, Christoph  and
      Fuchs, Susanne  and
      Stave, Matthew  and
      Seifart, Frank",
    editor = "Calzolari, Nicoletta  and
      B{\'e}chet, Fr{\'e}d{\'e}ric  and
      Blache, Philippe  and
      Choukri, Khalid  and
      Cieri, Christopher  and
      Declerck, Thierry  and
      Goggi, Sara  and
      Isahara, Hitoshi  and
      Maegaard, Bente  and
      Mariani, Joseph  and
      Mazo, H{\'e}l{\`e}ne  and
      Moreno, Asuncion  and
      Odijk, Jan  and
      Piperidis, Stelios",
    booktitle = "Proceedings of the Twelfth Language Resources and Evaluation Conference",
    month = may,
    year = "2020",
    address = "Marseille, France",
    publisher = "European Language Resources Association",
    url = "https://aclanthology.org/2020.lrec-1.324/",
    pages = "2657--2666",
    language = "eng",
    ISBN = "979-10-95546-34-4"
}

@inproceedings{zhu-etal-2024-taste,
    title = "The taste of {IPA}: Towards open-vocabulary keyword spotting and forced alignment in any language",
    author = "Zhu, Jian  and
      Yang, Changbing  and
      Samir, Farhan  and
      Islam, Jahurul",
    editor = "Duh, Kevin  and
      Gomez, Helena  and
      Bethard, Steven",
    booktitle = "Proceedings of the 2024 Conference of the North American Chapter of the Association for Computational Linguistics: Human Language Technologies (Volume 1: Long Papers)",
    month = jun,
    year = "2024",
    address = "Mexico City, Mexico",
    publisher = "Association for Computational Linguistics",
    url = "https://aclanthology.org/2024.naacl-long.43/",
    doi = "10.18653/v1/2024.naacl-long.43",
    pages = "750--772"
}

@article{pasad-etal-2024-self,
    title = "What Do Self-Supervised Speech Models Know About Words?",
    author = "Pasad, Ankita  and
      Chien, Chung-Ming  and
      Settle, Shane  and
      Livescu, Karen",
    journal = "Transactions of the Association for Computational Linguistics",
    volume = "12",
    year = "2024",
    address = "Cambridge, MA",
    publisher = "MIT Press",
    url = "https://aclanthology.org/2024.tacl-1.21/",
    doi = "10.1162/tacl_a_00656",
    pages = "372--391"
}

@inproceedings{meghanani-hain-2024-improving,
    title = "Improving Acoustic Word Embeddings through Correspondence Training of Self-supervised Speech Representations",
    author = "Meghanani, Amit  and
      Hain, Thomas",
    editor = "Graham, Yvette  and
      Purver, Matthew",
    booktitle = "Proceedings of the 18th Conference of the European Chapter of the Association for Computational Linguistics (Volume 1: Long Papers)",
    month = mar,
    year = "2024",
    address = "St. Julian{'}s, Malta",
    publisher = "Association for Computational Linguistics",
    url = "https://aclanthology.org/2024.eacl-long.118/",
    doi = "10.18653/v1/2024.eacl-long.118",
    pages = "1959--1967"
}

@inproceedings{mcauliffe2017montreal,
  title={Montreal forced aligner: Trainable text-speech alignment using kaldi.},
  author={McAuliffe, Michael and Socolof, Michaela and Mihuc, Sarah and Wagner, Michael and Sonderegger, Morgan},
  booktitle={Interspeech},
  volume={2017},
  pages={498--502},
  year={2017}
}

@article{shi2026qwen3,
  title={Qwen3-ASR Technical Report},
  author={Shi, Xian and Wang, Xiong and Guo, Zhifang and Wang, Yongqi and Zhang, Pei and Zhang, Xinyu and Guo, Zishan and Hao, Hongkun and Xi, Yu and Yang, Baosong and others},
  journal={arXiv preprint arXiv:2601.21337},
  year={2026}
}

@inproceedings{rastorgueva2023nemo,
  title={NeMo Forced Aligner and its application to word alignment for subtitle generation.},
  author={Rastorgueva, Elena and Lavrukhin, Vitaly and Ginsburg, Boris},
  booktitle={Interspeech},
  pages={5257--5258},
  year={2023}
}

@inproceedings{bain2023whisperx,
  title={WhisperX: Time-Accurate Speech Transcription of Long-Form Audio},
  author={Bain, Max and Huh, Jaesung and Han, Tengda and Zisserman, Andrew},
  booktitle={Interspeech},
  year={2023},
  publisher={ISCA}
}

@inproceedings{rousso2024tradition,
  title={Tradition or Innovation: A Comparison of Modern ASR Methods for Forced Alignment},
  author={Rousso, Rotem and Cohen, Eyal and Keshet, Joseph and Chodroff, Eleanor},
  booktitle={Proc. Interspeech 2024},
  pages={1525--1529},
  year={2024}
}

@inproceedings{kurzinger2020ctc,
  title={Ctc-segmentation of large corpora for german end-to-end speech recognition},
  author={K{\"u}rzinger, Ludwig and Winkelbauer, Dominik and Li, Lujun and Watzel, Tobias and Rigoll, Gerhard},
  booktitle={International Conference on Speech and Computer},
  pages={267--278},
  year={2020},
  organization={Springer}
}

@inproceedings{conneau2023fleurs,
  title={Fleurs: Few-shot learning evaluation of universal representations of speech},
  author={Conneau, Alexis and Ma, Min and Khanuja, Simran and Zhang, Yu and Axelrod, Vera and Dalmia, Siddharth and Riesa, Jason and Rivera, Clara and Bapna, Ankur},
  booktitle={2022 IEEE Spoken Language Technology Workshop (SLT)},
  pages={798--805},
  year={2023},
  organization={IEEE}
}

@inproceedings{conneau2021unsupervised,
  title={Unsupervised Cross-Lingual Representation Learning for Speech Recognition},
  author={Conneau, Alexis and Baevski, Alexei and Collobert, Ronan and Mohamed, Abdelrahman and Auli, Michael},
  booktitle={Proc. Interspeech 2021},
  pages={2426--2430},
  year={2021}
}

@article{baevski2020wav2vec,
  title={wav2vec 2.0: A framework for self-supervised learning of speech representations},
  author={Baevski, Alexei and Zhou, Yuhao and Mohamed, Abdelrahman and Auli, Michael},
  journal={Advances in neural information processing systems},
  volume={33},
  pages={12449--12460},
  year={2020}
}

@article{pratap2024scaling,
  title={Scaling speech technology to 1,000+ languages},
  author={Pratap, Vineel and Tjandra, Andros and Shi, Bowen and Tomasello, Paden and Babu, Arun and Kundu, Sayani and Elkahky, Ali and Ni, Zhaoheng and Vyas, Apoorv and Fazel-Zarandi, Maryam and others},
  journal={Journal of Machine Learning Research},
  volume={25},
  number={97},
  pages={1--52},
  year={2024}
}

@article{pitt2005buckeye,
  title={The Buckeye corpus of conversational speech: Labeling conventions and a test of transcriber reliability},
  author={Pitt, Mark A and Johnson, Keith and Hume, Elizabeth and Kiesling, Scott and Raymond, William},
  journal={Speech Communication},
  volume={45},
  number={1},
  pages={89--95},
  year={2005},
  publisher={Elsevier}
}

@misc{alekseevaetal2024,
  title = {Dictionary of Kina Rutul},
  author = {Alekseeva, Anastasia and Beklemishev, Nikita and Daniel, Michael and Dobrushina, Nina and Filatov, Konstantin and Ivanova, Anastasia and Maisak, Timur and Osorgin, Ivan},
  year = {2024},
  publisher = {Linguistic Convergence Laboratory, HSE University},
  address = {Moscow},
  url = {https://lingconlab.github.io/kina-rutul-dict/},
}

@misc{kibrik2007Archi,
    title = {Archi text corpus (1.0)},
    author = {Kibrik, Aleksandr E. and Kodzasov, Sandro V. and Olovyannikova, Irina P. and Samedov, Dzhalil S. and Daniel, Michael and Khoroshkina, Anna and Arkhipov, Alexandre},
    year = {2007},
    url = { https://doi.org/10.5281/zenodo.8247597}
}

@inproceedings{xu2022simple,
  title={Simple and Effective Zero-shot Cross-lingual Phoneme Recognition},
  author={Xu, Qiantong and Baevski, Alexei and Auli, Michael},
  booktitle={Proc. Interspeech 2022},
  pages={2113--2117},
  year={2022}
}

@Manual{XPF2021manual,
        author={Cohen Priva, Uriel and Strand, Emily and Yang, Shiying and Mizgerd, William and Creighton, Abigail and Bai, Justin and Mathew, Rebecca and Shao, Allison and Schuster, Jordan and Wiepert, Daniela},
        title =    {The Cross-linguistic Phonological Frequencies (XPF) Corpus manual},
        year =     {2021},
        note =     {Accessible online, \url{https://cohenpr-xpf.github.io/XPF/manual/xpf_manual.pdf}}
}

@article{kelley2024mason,
  title={The Mason-Alberta Phonetic Segmenter: a forced alignment system based on deep neural networks and interpolation},
  author={Kelley, Matthew C and Perry, Scott James and Tucker, Benjamin V},
  journal={Phonetica},
  volume={81},
  number={5},
  pages={451--508},
  year={2024},
  publisher={De Gruyter}
}

@inproceedings{gonzalez2018recursive,
  title={Recursive forced alignment: A test on a minority language},
  author={Gonzalez, Simon and Travis, Catherine and Grama, James and Barth, Danielle and Ananthanarayan, Sunkulp},
  booktitle={Proceedings of the 17th Australasian international conference on speech Science and technology},
  volume={145},
  pages={148},
  year={2018}
}

@article{hosom2009speaker,
  title={Speaker-independent phoneme alignment using transition-dependent states},
  author={Hosom, John-Paul},
  journal={Speech communication},
  volume={51},
  number={4},
  pages={352--368},
  year={2009},
  publisher={Elsevier}
}

@inproceedings{shi2022achieving,
  title={Achieving timestamp prediction while recognizing with non-autoregressive end-to-end asr model},
  author={Shi, Xian and Chen, Yanni and Zhang, Shiliang and Yan, Zhijie},
  booktitle={National Conference on Man-Machine Speech Communication},
  pages={89--100},
  year={2022},
  organization={Springer}
}

@inproceedings{de2024layer,
  title={A layer-wise analysis of Mandarin and English suprasegmentals in SSL speech models},
  author={de la Fuente, Anton and Jurafsky, Dan},
  booktitle={Proc. Interspeech 2024},
  pages={1290--1294},
  year={2024}
}

@inproceedings{omar2002evaluation,
  title={An evaluation of using mutual information for selection of acoustic-features representation of phonemes for speech recognition.},
  author={Omar, Mohamed Kamal and Chen, Ken and Hasegawa-Johnson, Mark and Brandman, Yigal},
  booktitle={Interspeech},
  pages={2129--2132},
  year={2002}
}

@article{cohen2021speaker,
  title={Speaker clustering quality estimation with logistic regression},
  author={Cohen, Yishai and Lapidot, Itshak},
  journal={Computer Speech \& Language},
  volume={65},
  pages={101139},
  year={2021},
  publisher={Elsevier}
}

@article{mridha2021u,
  title={U-vectors: Generating clusterable speaker embedding from unlabeled data},
  author={Mridha, Muhammad Firoz and Ohi, Abu Quwsar and Monowar, Muhammad Mostafa and Hamid, Md Abdul and Islam, Md Rashedul and Watanobe, Yutaka},
  journal={Applied Sciences},
  volume={11},
  number={21},
  pages={10079},
  year={2021},
  publisher={MDPI}
}

@article{tejedor2012comparison,
  title={Comparison of methods for language-dependent and language-independent query-by-example spoken term detection},
  author={Tejedor, Javier and Fap{\v{s}}o, Michal and Sz{\"o}ke, Igor and {\v{C}}ernock{\`y}, Jan “Honza” and Gr{\'e}zl, Franti{\v{s}}ek},
  journal={ACM Transactions on Information Systems (TOIS)},
  volume={30},
  number={3},
  pages={1--34},
  year={2012},
  publisher={ACM New York, NY, USA}
}

@article{sakoe1978dynamic,
  title={Dynamic programming algorithm optimization for spoken word recognition},
  author={Sakoe, Hiroaki and Chiba, Seibi},
  journal={IEEE transactions on acoustics, speech, and signal processing},
  volume={26},
  number={1},
  pages={43--49},
  year={1978},
  publisher={IEEE}
}

@article{jager2013phylogenetic,
  title={Phylogenetic inference from word lists using weighted alignment with empirically determined weights},
  author={J{\"a}ger, Gerhard},
  journal={Language Dynamics and Change},
  volume={3},
  number={2},
  pages={245--291},
  year={2013},
  publisher={Brill}
}

@article{jager2015support,
  title={Support for linguistic macrofamilies from weighted sequence alignment},
  author={J{\"a}ger, Gerhard},
  journal={Proceedings of the National Academy of Sciences},
  volume={112},
  number={41},
  pages={12752--12757},
  year={2015},
  publisher={National Academy of Sciences}
}

@article{scikit-learn,
  title={Scikit-learn: Machine Learning in {P}ython},
  author={Pedregosa, F. and Varoquaux, G. and Gramfort, A. and Michel, V.
          and Thirion, B. and Grisel, O. and Blondel, M. and Prettenhofer, P.
          and Weiss, R. and Dubourg, V. and Vanderplas, J. and Passos, A. and
          Cournapeau, D. and Brucher, M. and Perrot, M. and Duchesnay, E.},
  journal={Journal of Machine Learning Research},
  volume={12},
  pages={2825--2830},
  year={2011}
}

@inproceedings{houlsby2019parameter,
  title={Parameter-efficient transfer learning for NLP},
  author={Houlsby, Neil and Giurgiu, Andrei and Jastrzebski, Stanislaw and Morrone, Bruna and De Laroussilhe, Quentin and Gesmundo, Andrea and Attariyan, Mona and Gelly, Sylvain},
  booktitle={International conference on machine learning},
  pages={2790--2799},
  year={2019},
  organization={PMLR}
}

@inproceedings{loshchilov2017sgdr,
  title={SGDR: Stochastic Gradient Descent with Warm Restarts},
  author={Loshchilov, Ilya and Hutter, Frank},
  booktitle={International Conference on Learning Representations},
  year={2017}
}

@article{chodroff2018corpus,
  title={Corpus phonetics tutorial},
  author={Chodroff, Eleanor},
  journal={arXiv preprint arXiv:1811.05553},
  year={2018}
}

@misc{doreco-2-0,
  address   = {Lyon},
  author    = {Seifart, Frank and Paschen, Ludger and Stave, Matthew},
  howpublished = {Laboratoire Dynamique Du Langage (UMR5596, CNRS \& Université Lyon 2)},
  title     = {Language Documentation Reference Corpus ({DoReCo}) 2.0},
  url       = {https://doreco.huma-num.fr},
  doi       = {10.34847/nkl.7cbfq779},
  urldate   = {12/12/2024},
  year      = {2024}
}

@incollection{doreco-anal1239,
  address   = {Lyon},
  author    = {Ozerov, Pavel},
  booktitle = {Language Documentation Reference Corpus (DoReCo) 2.0},
  editor    = {Seifart, Frank and Paschen, Ludger and Stave, Matthew},
  publisher = {Laboratoire Dynamique Du Langage (UMR5596, CNRS \& Université Lyon 2)},
  title     = {Anal DoReCo dataset},
  url       = {https://doreco.huma-num.fr/languages/anal1239},
  doi       = {10.34847/nkl.0dbazp8m},
  urldate   = {12/12/2024},
  year      = {2024}
}

@incollection{doreco-apah1238,
  address   = {Lyon},
  author    = {Riesberg, Sonja},
  booktitle = {Language Documentation Reference Corpus (DoReCo) 2.0},
  editor    = {Seifart, Frank and Paschen, Ludger and Stave, Matthew},
  publisher = {Laboratoire Dynamique Du Langage (UMR5596, CNRS \& Université Lyon 2)},
  title     = {Yali (Apahapsili) DoReCo dataset},
  url       = {https://doreco.huma-num.fr/languages/apah1238},
  doi       = {10.34847/nkl.9d91nkq2},
  urldate   = {12/12/2024},
  year      = {2024}
}

@incollection{doreco-arap1274,
  address   = {Lyon},
  author    = {Cowell, Andrew},
  booktitle = {Language Documentation Reference Corpus (DoReCo) 2.0},
  editor    = {Seifart, Frank and Paschen, Ludger and Stave, Matthew},
  publisher = {Laboratoire Dynamique Du Langage (UMR5596, CNRS \& Université Lyon 2)},
  title     = {Arapaho DoReCo dataset},
  url       = {https://doreco.huma-num.fr/languages/arap1274},
  doi       = {10.34847/nkl.36f5r1b6},
  urldate   = {12/12/2024},
  year      = {2024}
}

@incollection{doreco-bain1259,
  address   = {Lyon},
  author    = {Cobbinah, Alexander Yao},
  booktitle = {Language Documentation Reference Corpus (DoReCo) 2.0},
  editor    = {Seifart, Frank and Paschen, Ludger and Stave, Matthew},
  publisher = {Laboratoire Dynamique Du Langage (UMR5596, CNRS \& Université Lyon 2)},
  title     = {Baïnounk Gubëeher DoReCo dataset},
  url       = {https://doreco.huma-num.fr/languages/bain1259},
  doi       = {10.34847/nkl.a332abw8},
  urldate   = {12/12/2024},
  year      = {2024}
}

@incollection{doreco-beja1238,
  address   = {Lyon},
  author    = {Vanhove, Martine},
  booktitle = {Language Documentation Reference Corpus (DoReCo) 2.0},
  editor    = {Seifart, Frank and Paschen, Ludger and Stave, Matthew},
  publisher = {Laboratoire Dynamique Du Langage (UMR5596, CNRS \& Université Lyon 2)},
  title     = {Beja DoReCo dataset},
  url       = {https://doreco.huma-num.fr/languages/beja1238},
  doi       = {10.34847/nkl.edd011t1},
  urldate   = {12/12/2024},
  year      = {2024}
}

@incollection{doreco-bora1263,
  address   = {Lyon},
  author    = {Seifart, Frank},
  booktitle = {Language Documentation Reference Corpus (DoReCo) 2.0},
  editor    = {Seifart, Frank and Paschen, Ludger and Stave, Matthew},
  publisher = {Laboratoire Dynamique Du Langage (UMR5596, CNRS \& Université Lyon 2)},
  title     = {Bora DoReCo dataset},
  url       = {https://doreco.huma-num.fr/languages/bora1263},
  doi       = {10.34847/nkl.6eaf5laq},
  urldate   = {12/12/2024},
  year      = {2024}
}

@incollection{doreco-cabe1245,
  address   = {Lyon},
  author    = {Quesada, Juan Diego and Skopeteas, Stavros and Pasamonik, Carolina and Brokmann, Carolin and Fischer, Florian},
  booktitle = {Language Documentation Reference Corpus (DoReCo) 2.0},
  editor    = {Seifart, Frank and Paschen, Ludger and Stave, Matthew},
  publisher = {Laboratoire Dynamique Du Langage (UMR5596, CNRS \& Université Lyon 2)},
  title     = {Cabécar DoReCo dataset},
  url       = {https://doreco.huma-num.fr/languages/cabe1245},
  doi       = {10.34847/nkl.ebc4ra22},
  urldate   = {12/12/2024},
  year      = {2024}
}

@incollection{doreco-cash1254,
  address   = {Lyon},
  author    = {Reiter, Sabine},
  booktitle = {Language Documentation Reference Corpus (DoReCo) 2.0},
  editor    = {Seifart, Frank and Paschen, Ludger and Stave, Matthew},
  publisher = {Laboratoire Dynamique Du Langage (UMR5596, CNRS \& Université Lyon 2)},
  title     = {Cashinahua DoReCo dataset},
  url       = {https://doreco.huma-num.fr/languages/cash1254},
  doi       = {10.34847/nkl.a8f9q2f1},
  urldate   = {12/12/2024},
  year      = {2024}
}

@article{hsu2021hubert,
  title={Hubert: Self-supervised speech representation learning by masked prediction of hidden units},
  author={Hsu, Wei-Ning and Bolte, Benjamin and Tsai, Yao-Hung Hubert and Lakhotia, Kushal and Salakhutdinov, Ruslan and Mohamed, Abdelrahman},
  journal={IEEE/ACM transactions on audio, speech, and language processing},
  volume={29},
  pages={3451--3460},
  year={2021},
  publisher={IEEE}
}

\newpage
\appendix
\section*{Appendix}

\section{Implementation details}
\label{app:impl}

Alignment evaluations were conducted using a single NVIDIA RTX 2080 GPU (11GB VRAM) on an Intel Xeon Gold 6140 2.30GHz (80GB RAM) machine with batch size 2.

For PCMI evaluation, we sample up to 50 utterances per language for phoneme-cluster estimation and limit evaluation to at most 10\,000 acoustic frames. Frame sampling is performed without replacement, where phoneme labels are sampled proportionally to the square root of their frequency in order to reduce dominance from highly frequent labels. Clustering is performed using MiniBatch K-means with batch size 1000 using Scikit-learn \citep{scikit-learn}.

For WACS evaluation, we sample up to 200 utterances and up to 200 unique word forms per language, although the effective number is additionally constrained by the requirement that each occurrence span at least four embedding frames. To maximize the sampling space for each word form, forms with a larger number of occurrences are preferentially selected by sorting. Further, only word forms appearing at least three times are considered, ensuring that positive similarity estimates are computed from multiple repeated realizations. To avoid dominance by highly frequent word forms, at most 10 positive pairs (from $\Sigma^+_f$) are sampled for each word form $f$. For every sampled word form $f$, five randomly sampled negative pairs (from $\Sigma^-_f$) are additionally considered.

The subsampling strategies used for both PCMI and WACS substantially improve throughput while maintaining stable aggregate estimates with relatively small variances across evaluations. Runtime statistics, vocabulary sizes, and pair statistics are reported in Appendix Table~\ref{tab:multilingual-stats}. Evaluating the complete multilingual FLEURS benchmark (85 languages) requires approximately 40 minutes per embedding model, corresponding to roughly 30 seconds per language under the defined sampling configuration. The code for reproducing the experiments is available at \url{https://github.com/mahesh-ak/MFA}.

\section{WACS Similarity Computation}
\label{app:wacs_details}

For WACS, each word occurrence is represented as a sequence of frame-level speech representations
$\mathbf{e}_w=\{\mathbf{emb}_1,\ldots,\mathbf{emb}_{|w|}\}$,
where $\mathbf{emb}_i \in \mathbb{R}^d$ denotes the embedding extracted from the selected SSL-speech model layer. Given two word occurrences $w_1$ and $w_2$, we compute pairwise frame similarities between frames $i$ and $j$ using cosine similarity:
\[
s(i,j) = \frac{\mathbf{emb}_i^\top \mathbf{emb}_j}{||\mathbf{emb}_i||\cdot||\mathbf{emb}_j||}
\]

Since different realizations of the same word may vary in speaking rate and duration, the resulting embedding sequences generally have different lengths. We therefore apply dynamic time warping (DTW) \citep{sakoe1978dynamic} to find a monotonic alignment path through the frame-similarity matrix. The DTW score is computed as the average cosine similarity along the optimal path:

\[
\mathrm{dtw}(\mathbf{e}_{w_1},\mathbf{e}_{w_2})
=
\frac{1}{|\pi^\ast|}
\sum_{(i,j)\in\pi^\ast}
s(i,j),
\]

where $\pi^\ast$ denotes the optimal DTW alignment path and $|\pi^\ast|$ its length. This procedure yields a duration-invariant similarity measure between two word realizations while preserving the acoustic structure encoded by the speech representations.

\section{PCMI Under Label Collapse}
\label{app:pcmi-collapse}

A potential concern is whether degenerate collapse of phoneme labels could artificially inflate PCMI through entropy reduction. We show that this cannot occur. Given:
\[
\mathrm{PCMI}=\frac{I(P;C)}{\sqrt{H(P)H(C)}},
\]
where $P$ denotes phoneme labels, $C$ denotes representation clusters and $H(\cdot)$ denotes Shannon entropy for phoneme probabilities $\{p_i\}$:
\[ H(P) = -\sum_i p_i \log p_i\]
Since entropy is non-negative, the condition $H(P \mid C) \geq 0$ yields:
\[I(P;C) = H(P) - H(P \mid C) \le H(P),\]
we obtain,
\[
\mathrm{PCMI}\le\frac{H(P)}{\sqrt{H(P)H(C)}}=\sqrt{\frac{H(P)}{H(C)}}.
\]

Assuming $H(C)>0$, which holds empirically since representation clusters remain non-degenerate across experiments, we have
\[
\mathrm{PCMI}\to 0\quad \text{as} \quad H(P)\to0.
\]
Therefore, complete collapse of the phoneme inventory cannot artificially inflate PCMI. 

We note that this argument guards against degenerate collapse; small-scale systematic merging of acoustically similar phoneme categories may still occur and can affect PCMI as demonstrated in \S\ref{subsec:perturb}.

\section{Phoneme Models Training}
\label{app:ipa-train}

We finetune CTC-based phoneme recognition models MMS-300M-IPA/MMS-300M-DORECO and Wav2Vec2-IPA respectively using MMS-300M \citep{pratap2024scaling} and Wav2Vec2-Phoneme \citep{xu2022simple} (both of size $\sim$ 300M parameters) on the multilingual IPA transcriptions derived from FLEURS as described in \S\ref{subsec:fleurs}. Following \citet{pratap2024scaling}, language-specific adapter modules \citep{houlsby2019parameter} are used during finetuning.

Training uses an effective batch size of 16, with per-device batch size 2 on two NVIDIA RTX 2080 GPUs (11GB VRAM each), gradient accumulation steps 4, and fp16 precision. Models are optimized using AdamW with learning rate $10^{-5}$ and cosine learning-rate scheduling with warmup ratio 0.01 \citep{loshchilov2017sgdr}. We first train the backbone models without adapters for 20\,000 steps (approximately three epochs), followed by an additional 500 steps per language with adapters enabled and learning-rate restart. ASR performances of the resulting phoneme recognition models are reported in Appendix Table~\ref{tab:multilingual_asr} and \ref{tab:doreco_asr}. The training time per model is about 20hrs. MMS-300M-DORECO is finetuned starting from MMS-300M-IPA for 8000 steps before training the language specific adapters. The models with best performances are available at \url{https://hf.co/mahesh27/mms-300m-ipa-fleurs} and \url{https://hf.co/mahesh27/mms-300m-xsampa-doreco}

\paragraph{CTC Forced Aligner} Forced alignment is performed using CTC segmentation \citep{kurzinger2020ctc} at both phoneme and word levels, followed by additional sanity checks and postprocessing to produce valid TextGrid annotations. Alignment inference is performed on a single GPU. 

\paragraph{Silence Correction} For the silence-absorbing variant (-SIL), we apply a simple energy-based silence trimming procedure. First, an absolute-amplitude envelope is computed and smoothed using a moving-average window of 400 samples (approximately 25 ms at 16 kHz). A global silence threshold is then defined as 5\% of the 95th percentile of the smoothed envelope. Within each aligned word interval, leading and trailing regions whose energy falls below this threshold are removed by locating the first and last frames exceeding the threshold. Intervals shorter than a minimum word duration (40ms) are left unchanged. The resulting word-level boundary adjustments are subsequently propagated to the first and last phonemes overlapping each word while enforcing minimum phoneme durations (20ms). This procedure removes extended silent regions absorbed into neighboring words without otherwise modifying the internal phoneme segmentation.

\paragraph{MFA Training} For MFA \citep{mcauliffe2017montreal}, we train multilingual acoustic models using the same IPA-transcribed FLEURS training data on an Intel Core i9-14900K CPU system with 64GB RAM. Training each language required approximately 1.5--2 hours. The training configuration broadly follows \citet{ahn-chodroff-2022-voxcommunis}, using MFCC features with 10\,ms frame shift and successive stages of monophone training, triphone training, Linear Discriminant Analysis (LDA), and two stages of Speaker Adaptation Training (SAT). The corresponding Gaussian mixture sizes are 1000, 10000, 15000, 15000, and 20000 respectively. The triphone, LDA, and SAT stages use 2000, 2500, 2500, and 3000 phonetic decision-tree leaves respectively \citep{chodroff2018corpus}. Due to inconsistent speaker metadata in FLEURS, MFA training is performed in single-speaker mode despite the dataset containing balanced male and female speakers.

\paragraph{Dataset splits} For DoReCo, the recordings are segmented into manually annotated utterance-level chunks. Only chunks containing at least two words are retained. For each language, the retained chunks are randomly partitioned into training, development, and test sets using a 0.7/0.1/0.2 split. For FLEURS, the original predefined train/dev/test splits are used without modification. All models are trained on the training splits, while all alignment evaluations are performed exclusively on the corresponding test splits. The statistics of datasets are provided in Tables \ref{tab:language_info} and \ref{tab:doreco-languages}.

\onecolumn
\newpage
\section{Supplementary Figures}

\begin{figure}[h!]
    \centering
    \includegraphics[width=0.7\linewidth]{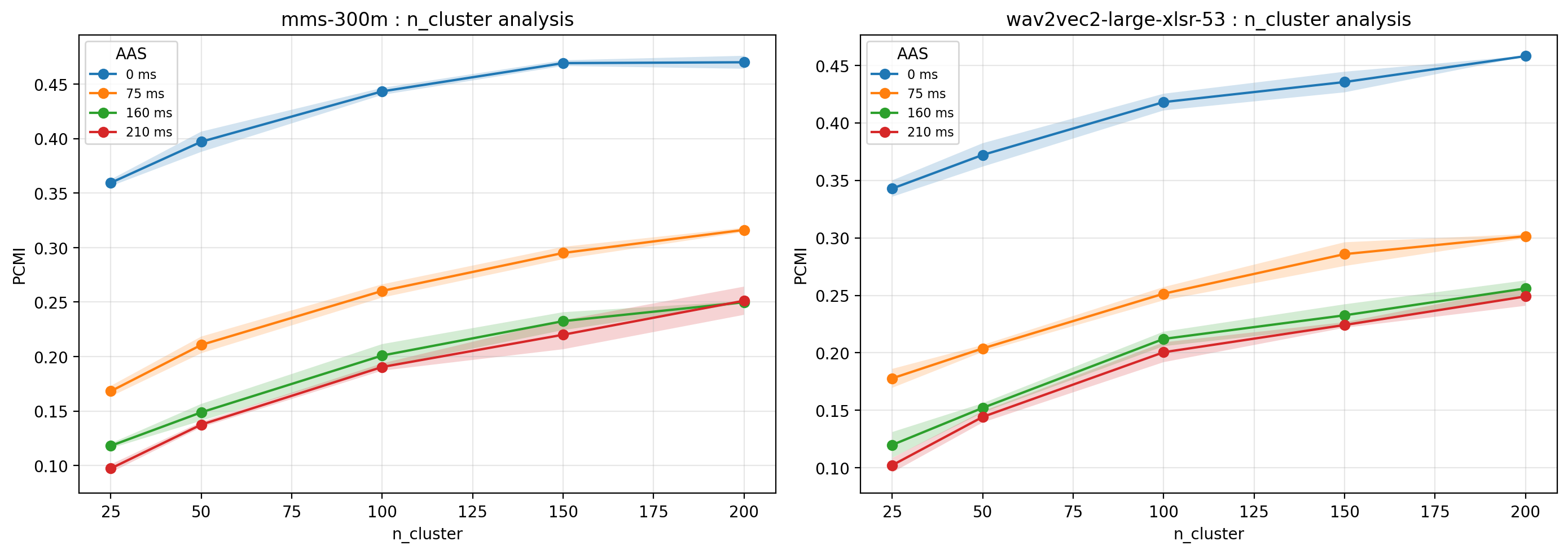}
   \caption{Behavior of PCMI under varying numbers of K-means clusters and increasing perturbation severity using MMS (left) and XLSR (right) representations. The metric resolution, measured as the gap between clean and severely perturbed alignments, remains stable over cluster counts $25$--$100$.}
    \label{fig:clusters}
\end{figure}

\begin{figure*}[h!]
    \centering
    \includegraphics[width=\linewidth]{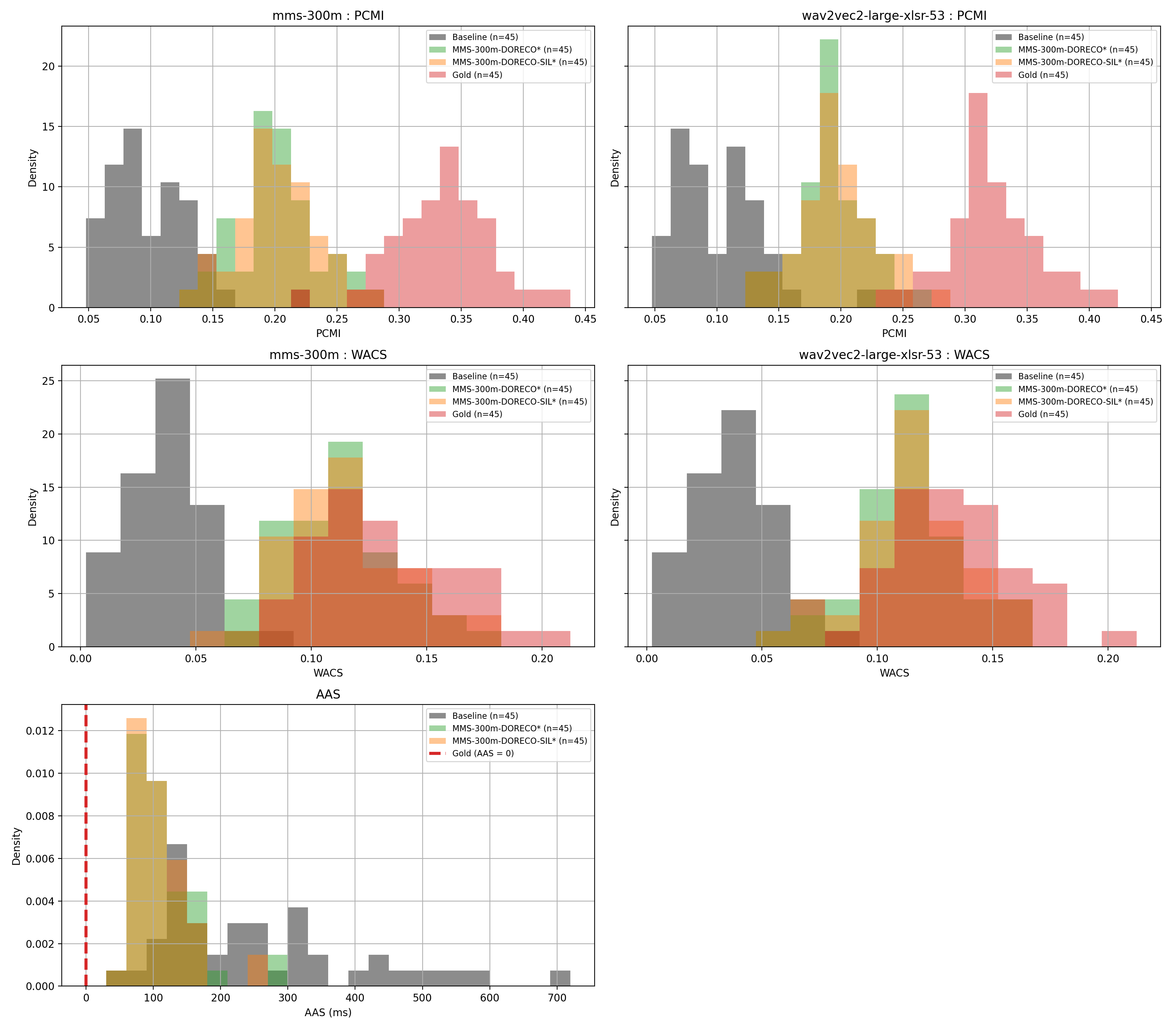}
    \caption{
    Distribution of PCMI (top row), WACS (middle) and AAS (bottom) across multilingual alignments on DoReCo corpus. The left column uses MMS representations while the right uses XLSR representations. 
    \\ * Trained in this work.
    }
    \label{fig:multilingual_doreco}
\end{figure*}

\newpage
\section{Supplementary Tables}

\begingroup
\small
\setlength{\tabcolsep}{2.8pt}
\renewcommand{\arraystretch}{0.92}
\begin{longtable}{lllllllllll}
\caption{Languages from FLEURS \citep{conneau2023fleurs} used in multilingual evaluation.}\label{tab:language_info}\\
\toprule
\textbf{Language} & \textbf{ISO-1} & \textbf{ISO-3} & \textbf{Family} & \textbf{Subfamily} & \textbf{Division} & \textbf{Train} & \textbf{Dev} & \textbf{Test} & \textbf{G2P} \\
\midrule
\endfirsthead
\toprule
\textbf{Language} & \textbf{ISO-1} & \textbf{ISO-3} & \textbf{Family} & \textbf{Subfamily} & \textbf{Division} & \textbf{Train} & \textbf{Dev} & \textbf{Test} & \textbf{G2P} \\
\midrule
\endhead
\bottomrule
\endfoot
Afrikaans & af & afr & Indo-European & Germanic & West & 1032 & 198 & 264 & epitran \\
Amharic & am & amh & Afroasiatic & Semitic & South & 3163 & 223 & 516 & epitran \\
Arabic & ar & ara & Afroasiatic & Semitic & Central & 2104 & 295 & 428 & epitran \\
Asturian & - & ast & Indo-European & Romance & West & 2511 & 398 & 946 & xpf \\
Azerbaijani & az & aze & Turkic & Oghuz & - & 2665 & 400 & 923 & epitran \\
Belarusian & be & bel & Indo-European & Slavic & East & 2433 & 408 & 967 & xpf \\
Bulgarian & bg & bul & Indo-European & Slavic & South & 2973 & 395 & 658 & xpf \\
Bengali & bn & ben & Indo-European & Indo-Aryan & East & 3006 & 402 & 920 & epitran \\
\end{longtable}
\endgroup
\begingroup
\scriptsize
\setlength{\tabcolsep}{3pt}
\renewcommand{\arraystretch}{0.92}
\begin{longtable}{lllllllll}
\caption{DoReCo \citep{doreco-2-0} languages used for multilingual validation.}\label{tab:doreco-languages}\\
\toprule
\textbf{Language} & \textbf{ISO-3} & \textbf{Family} & \textbf{Subfamily} & \textbf{Division} & \textbf{Train} & \textbf{Dev} & \textbf{Test} & \textbf{Source} \\
\midrule
\endfirsthead
\toprule
\textbf{Language} & \textbf{ISO-3} & \textbf{Family} & \textbf{Subfamily} & \textbf{Division} & \textbf{Train} & \textbf{Dev} & \textbf{Test} & \textbf{Source} \\
\midrule
\endhead
\bottomrule
\endfoot
Anal & anm & Sino-Tibetan & Tibeto-Burman & Kuki-Chin & 1588 & 227 & 453 & \citet{doreco-anal1239} \\
Yali (Apahapsili) & na & Trans-New Guinea & Dani & - & 1057 & 151 & 302 & \citet{doreco-apah1238} \\
Arapaho & arp & Algic & Algonquian & - & 782 & 112 & 223 & \citet{doreco-arap1274} \\
Baïnounk Gubëeher & bab & Niger-Congo & West-Atlantic & - & 1315 & 188 & 376 & \citet{doreco-bain1259} \\
Beja & bej & Afroasiatic & Cushitic & North & 3172 & 453 & 906 & \citet{doreco-beja1238} \\
Bora & boa & Boran & - & - & 733 & 105 & 209 & \citet{doreco-bora1263} \\
Cabécar & cjp & Chibchan & Isthmic & - & 801 & 114 & 229 & \citet{doreco-cabe1245} \\
Cashinahua & cbs & Pano-Tacanan & Panoan & - & 1185 & 169 & 339 & \citet{doreco-cash1254} \\
\end{longtable}
\endgroup
\begingroup
\small
\setlength{\tabcolsep}{3pt}
\renewcommand{\arraystretch}{0.92}
\begin{longtable}{lp{0.8cm}cccc}
\caption{Language-wise multilingual ASR performance on FLEURS. $N$ denotes sizes of test splits. MMS-IPA* and W2V2-IPA* are same as MMS-300M-IPA* and Wav2Vec2-IPA* respectively. * Trained in this work.}\label{tab:multilingual_asr}\\
\toprule
\multirow{2}{*}{\textbf{Language}} & \multirow{2}{*}{\textbf{N}} & \multicolumn{2}{c}{\textbf{WER} $\downarrow$} & \multicolumn{2}{c}{\textbf{CER} $\downarrow$} \\
\cmidrule(lr){3-4}\cmidrule(lr){5-6}
& & \textbf{MMS-IPA*} & \textbf{W2V2-IPA*} & \textbf{MMS-IPA*} & \textbf{W2V2-IPA*} \\
\midrule
\endfirsthead
\toprule
\multirow{2}{*}{\textbf{Language}} & \multirow{2}{*}{\textbf{N}} & \multicolumn{2}{c}{\textbf{WER} $\downarrow$} & \multicolumn{2}{c}{\textbf{CER} $\downarrow$} \\
\cmidrule(lr){3-4}\cmidrule(lr){5-6}
& & \textbf{MMS-IPA*} & \textbf{W2V2-IPA*} & \textbf{MMS-IPA*} & \textbf{W2V2-IPA*} \\
\midrule
\endhead
\bottomrule
\endfoot
Afrikaans & 264 & \textbf{0.576} & 0.852 & \textbf{0.190} & 0.323 \\
Amharic & 516 & \textbf{0.638} & 0.866 & \textbf{0.124} & 0.208 \\
Arabic & 428 & \textbf{0.526} & 0.927 & \textbf{0.141} & 0.308 \\
Asturian & 946 & \textbf{0.370} & 0.694 & \textbf{0.086} & 0.168 \\
Azerbaijani & 923 & \textbf{0.509} & 0.839 & \textbf{0.113} & 0.223 \\
Belarusian & 967 & \textbf{0.378} & 0.805 & \textbf{0.079} & 0.176 \\
Bulgarian & 658 & \textbf{0.330} & 0.794 & \textbf{0.077} & 0.207 \\
Bengali & 920 & \textbf{0.649} & 0.849 & \textbf{0.149} & 0.248 \\
\end{longtable}
\endgroup
\begin{table*}[h!]
\centering
\small
\setlength{\tabcolsep}{3pt}
\renewcommand{\arraystretch}{0.92}
\caption{Language-wise ASR performance on DoReCo. Scores are reported for MMS-300M-DORECO*. $N$ denotes sizes of test splits}
\begin{tabular}{lccc}
\toprule
\textbf{Language} & \textbf{N} & \textbf{WER} $\downarrow$ & \textbf{CER} $\downarrow$ \\
\midrule
Anal & 453 & 0.979 & 0.454 \\
Arapaho & 223 & 1.007 & 0.630 \\
Asimjeeg Datooga & 325 & 0.989 & 0.356 \\
Baïnounk Gubëeher & 376 & 0.957 & 0.366 \\
Beja & 906 & 0.975 & 0.322 \\
Bora & 209 & 0.998 & 0.399 \\
Cabécar & 229 & 0.991 & 0.460 \\
Cashinahua & 339 & 0.991 & 0.342 \\
Daakie & 263 & 0.894 & 0.375 \\
Dalabon & 83 & 0.968 & 0.313 \\
Dolgan & 228 & 0.909 & 0.357 \\
English (Southern England) & 123 & 0.927 & 0.493 \\
Evenki & 374 & 0.977 & 0.405 \\
Fanbyak & 303 & 0.830 & 0.291 \\
French (Swiss) & 349 & 0.743 & 0.308 \\
Gorwaa & 285 & 0.930 & 0.386 \\
Jahai & 239 & 0.856 & 0.322 \\
Jejuan & 88 & 0.974 & 0.443 \\
Kakabe & 191 & 0.805 & 0.260 \\
Kamas & 363 & 0.897 & 0.323 \\
Komnzo & 436 & 0.969 & 0.344 \\
Light Warlpiri & 198 & 0.901 & 0.405 \\
Mojeño Trinitario & 179 & 0.744 & 0.208 \\
\bottomrule
\end{tabular}\hspace*{1cm}
\begin{tabular}{lccc}
\toprule
\textbf{Language} & \textbf{N} & \textbf{WER} $\downarrow$ & \textbf{CER} $\downarrow$ \\
\midrule
Nafsan (South Efate) & 155 & 0.864 & 0.316 \\
Nisvai & 339 & 0.928 & 0.309 \\
Northern Alta & 378 & 0.815 & 0.312 \\
Northern Kurdish (Kurmanji) & 167 & 0.991 & 0.328 \\
Nllng & 311 & 0.999 & 0.645 \\
Pnar & 63 & 0.810 & 0.368 \\
Resígaro & 269 & 0.954 & 0.273 \\
Ruuli & 233 & 0.979 & 0.321 \\
Sadu & 296 & 0.956 & 0.428 \\
Sanzhi Dargwa & 90 & 0.949 & 0.394 \\
Savosavo & 179 & 0.761 & 0.210 \\
Svan & 142 & 0.968 & 0.334 \\
Sümi & 231 & 0.977 & 0.470 \\
Tabasaran & 125 & 0.883 & 0.331 \\
Teop & 379 & 0.907 & 0.295 \\
Texistepec Popoluca & 471 & 1.002 & 0.624 \\
Urum & 254 & 0.919 & 0.391 \\
Vera'a & 269 & 0.855 & 0.352 \\
Warlpiri & 335 & 0.794 & 0.198 \\
Yali (Apahapsili) & 302 & 0.936 & 0.415 \\
Yongning Na & 133 & 0.990 & 0.406 \\
Yucatec Maya & 225 & 0.934 & 0.371 \\
 & & & \\
\bottomrule
\end{tabular}
\label{tab:doreco_asr}
\end{table*}
\begingroup
\small
\setlength{\tabcolsep}{3pt}
\renewcommand{\arraystretch}{0.92}
\begin{longtable}{lcccccccc}
\caption{Language-wise multilingual PCMI scores on FLEURS. Higher values are better. MMS-IPA* and W2V2-IPA* are same as MMS-300M-IPA* and Wav2Vec2-IPA* respectively.  * Trained in this work.}\label{tab:pcmi_table}\\
\toprule
\multirow{2}{*}{\textbf{Language}} & \multicolumn{4}{c}{\textbf{MMS}} & \multicolumn{4}{c}{\textbf{XLSR}} \\
\cmidrule(lr){2-5}\cmidrule(lr){6-9}
& \textbf{Base}& \textbf{MFA}& \textbf{MMS-IPA*}& \textbf{W2V2-IPA*}& \textbf{Base}& \textbf{MFA}& \textbf{MMS-IPA*}& \textbf{W2V2-IPA*}\\
\midrule
\endfirsthead
\toprule
\multirow{2}{*}{\textbf{Language}} & \multicolumn{4}{c}{\textbf{MMS}} & \multicolumn{4}{c}{\textbf{XLSR}} \\
\cmidrule(lr){2-5}\cmidrule(lr){6-9}
& \textbf{Base}& \textbf{MFA}& \textbf{MMS-IPA*}& \textbf{W2V2-IPA*}& \textbf{Base}& \textbf{MFA}& \textbf{MMS-IPA*}& \textbf{W2V2-IPA*}\\
\midrule
\endhead
\bottomrule
\endfoot
Afrikaans & 0.093 & 0.112 & 0.221 & \textbf{0.226} & 0.101 & 0.105 & \textbf{0.218} & 0.215 \\
Amharic & 0.103 & 0.122 & \textbf{0.258} & 0.251 & 0.108 & 0.120 & \textbf{0.249} & 0.234 \\
Arabic & 0.114 & \textbf{0.316} & 0.254 & 0.254 & 0.107 & \textbf{0.281} & 0.244 & 0.232 \\
Asturian & 0.091 & 0.117 & \textbf{0.223} & 0.219 & 0.096 & 0.106 & 0.199 & \textbf{0.203} \\
Azerbaijani & 0.071 & 0.124 & 0.228 & \textbf{0.243} & 0.076 & 0.126 & \textbf{0.227} & 0.220 \\
Belarusian & 0.075 & \textbf{0.352} & 0.247 & 0.231 & 0.078 & \textbf{0.335} & 0.237 & 0.233 \\
Bulgarian & 0.099 & 0.082 & 0.212 & \textbf{0.213} & 0.097 & 0.080 & 0.184 & \textbf{0.193} \\
Bengali & 0.081 & \textbf{0.273} & 0.241 & 0.224 & 0.085 & \textbf{0.250} & 0.224 & 0.225 \\
\end{longtable}
\endgroup
\begingroup
\scriptsize
\setlength{\tabcolsep}{3pt}
\renewcommand{\arraystretch}{0.92}
\begin{longtable}{lcccccccccc}
\caption{Language-wise multilingual WACS scores on FLEURS. Higher values are better. MMS-IPA* and W2V2-IPA* are same as MMS-300M-IPA* and Wav2Vec2-IPA* respectively. * Trained in this work.}\label{tab:wacs_table}\\
\toprule
\multirow{2}{*}{\textbf{Language}} & \multicolumn{5}{c}{\textbf{MMS}} & \multicolumn{5}{c}{\textbf{XLSR}} \\
\cmidrule(lr){2-6}\cmidrule(lr){7-11}
& \textbf{Base}& \textbf{MFA}& \textbf{MMS-IPA*}& \textbf{W2V2-IPA*}& \textbf{Qwen3-FA}& \textbf{Base}& \textbf{MFA}& \textbf{MMS-IPA*}& \textbf{W2V2-IPA*}& \textbf{Qwen3-FA}\\
\midrule
\endfirsthead
\toprule
\multirow{2}{*}{\textbf{Language}} & \multicolumn{5}{c}{\textbf{MMS}} & \multicolumn{5}{c}{\textbf{XLSR}} \\
\cmidrule(lr){2-6}\cmidrule(lr){7-11}
& \textbf{Base}& \textbf{MFA}& \textbf{MMS-IPA*}& \textbf{W2V2-IPA*}& \textbf{Qwen3-FA}& \textbf{Base}& \textbf{MFA}& \textbf{MMS-IPA*}& \textbf{W2V2-IPA*}& \textbf{Qwen3-FA}\\
\midrule
\endhead
\bottomrule
\endfoot
Afrikaans & 0.008 & 0.041 & \textbf{0.181} & 0.173 & - & 0.008 & 0.025 & \textbf{0.161} & 0.159 & - \\
Amharic & 0.056 & 0.044 & 0.219 & \textbf{0.225} & - & 0.053 & 0.044 & \textbf{0.198} & 0.198 & - \\
Arabic & 0.041 & \textbf{0.208} & 0.207 & 0.206 & - & 0.034 & \textbf{0.185} & 0.177 & 0.178 & - \\
Asturian & 0.032 & 0.052 & \textbf{0.186} & 0.182 & - & 0.023 & 0.053 & 0.172 & \textbf{0.172} & - \\
Azerbaijani & 0.026 & 0.031 & 0.220 & \textbf{0.220} & - & 0.032 & 0.033 & \textbf{0.195} & 0.192 & - \\
Belarusian & 0.005 & \textbf{0.231} & 0.195 & 0.190 & - & 0.004 & \textbf{0.200} & 0.168 & 0.164 & - \\
Bulgarian & 0.039 & 0.024 & \textbf{0.209} & 0.203 & - & 0.031 & 0.027 & \textbf{0.176} & 0.169 & - \\
Bengali & 0.026 & 0.192 & \textbf{0.204} & 0.196 & - & 0.024 & 0.164 & \textbf{0.174} & 0.170 & - \\
\end{longtable}
\endgroup
\begingroup
\scriptsize
\setlength{\tabcolsep}{3pt}
\renewcommand{\arraystretch}{0.92}
\begin{longtable}{lcccccccc}
\caption{Language-wise multilingual PCMI scores on DoReCo. Higher values are better. * Trained in this work.}\label{tab:pcmi_table_doreco}\\
\toprule
\multirow{2}{*}{\textbf{Language}} & \multicolumn{4}{c}{\textbf{MMS}} & \multicolumn{4}{c}{\textbf{XLSR}} \\
\cmidrule(lr){2-5}\cmidrule(lr){6-9}
& \textbf{Base}& \textbf{MMS-DORECO*}& \textbf{MMS-DORECO*-SIL}& \textbf{Gold}& \textbf{Base}& \textbf{MMS-DORECO*}& \textbf{MMS-DORECO*-SIL}& \textbf{Gold}\\
\midrule
\endfirsthead
\toprule
\multirow{2}{*}{\textbf{Language}} & \multicolumn{4}{c}{\textbf{MMS}} & \multicolumn{4}{c}{\textbf{XLSR}} \\
\cmidrule(lr){2-5}\cmidrule(lr){6-9}
& \textbf{Base}& \textbf{MMS-DORECO*}& \textbf{MMS-DORECO*-SIL}& \textbf{Gold}& \textbf{Base}& \textbf{MMS-DORECO*}& \textbf{MMS-DORECO*-SIL}& \textbf{Gold}\\
\midrule
\endhead
\bottomrule
\endfoot
Anal & 0.133 & 0.209 & 0.206 & \textbf{0.330} & 0.122 & 0.197 & 0.195 & \textbf{0.316} \\
Yali (Apahapsili) & 0.127 & 0.196 & 0.193 & \textbf{0.263} & 0.125 & 0.184 & 0.184 & \textbf{0.248} \\
Arapaho & 0.075 & 0.192 & 0.210 & \textbf{0.287} & 0.072 & 0.193 & 0.204 & \textbf{0.292} \\
Baïnounk Gubëeher & 0.097 & 0.185 & 0.194 & \textbf{0.301} & 0.098 & 0.182 & 0.193 & \textbf{0.308} \\
Beja & 0.228 & 0.284 & 0.279 & \textbf{0.386} & 0.217 & 0.270 & 0.278 & \textbf{0.363} \\
Bora & 0.061 & 0.193 & 0.186 & \textbf{0.325} & 0.069 & 0.191 & 0.193 & \textbf{0.317} \\
Cabécar & 0.073 & 0.135 & 0.138 & \textbf{0.338} & 0.070 & 0.137 & 0.125 & \textbf{0.333} \\
Cashinahua & 0.084 & 0.138 & 0.139 & \textbf{0.335} & 0.083 & 0.138 & 0.138 & \textbf{0.313} \\
\end{longtable}
\endgroup
\begingroup
\scriptsize
\setlength{\tabcolsep}{3pt}
\renewcommand{\arraystretch}{0.92}
\begin{longtable}{lcccccccc}
\caption{Language-wise multilingual WACS scores on DoReCo. Higher values are better. * Trained in this work.}\label{tab:wacs_table_doreco}\\
\toprule
\multirow{2}{*}{\textbf{Language}} & \multicolumn{4}{c}{\textbf{MMS}} & \multicolumn{4}{c}{\textbf{XLSR}} \\
\cmidrule(lr){2-5}\cmidrule(lr){6-9}
& \textbf{Base}& \textbf{MMS-DORECO*}& \textbf{MMS-DORECO*-SIL}& \textbf{Gold}& \textbf{Base}& \textbf{MMS-DORECO*}& \textbf{MMS-DORECO*-SIL}& \textbf{Gold}\\
\midrule
\endfirsthead
\toprule
\multirow{2}{*}{\textbf{Language}} & \multicolumn{4}{c}{\textbf{MMS}} & \multicolumn{4}{c}{\textbf{XLSR}} \\
\cmidrule(lr){2-5}\cmidrule(lr){6-9}
& \textbf{Base}& \textbf{MMS-DORECO*}& \textbf{MMS-DORECO*-SIL}& \textbf{Gold}& \textbf{Base}& \textbf{MMS-DORECO*}& \textbf{MMS-DORECO*-SIL}& \textbf{Gold}\\
\midrule
\endhead
\bottomrule
\endfoot
Anal & 0.042 & 0.091 & 0.097 & \textbf{0.104} & 0.046 & 0.096 & 0.095 & \textbf{0.105} \\
Yali (Apahapsili) & 0.049 & 0.084 & \textbf{0.085} & 0.083 & 0.053 & 0.095 & 0.101 & \textbf{0.111} \\
Arapaho & 0.071 & 0.100 & 0.116 & \textbf{0.136} & 0.072 & 0.115 & 0.123 & \textbf{0.151} \\
Baïnounk Gubëeher & 0.037 & 0.092 & 0.102 & \textbf{0.114} & 0.038 & 0.093 & 0.097 & \textbf{0.114} \\
Beja & 0.084 & 0.104 & 0.092 & \textbf{0.111} & 0.083 & 0.118 & 0.119 & \textbf{0.127} \\
Bora & 0.043 & 0.118 & 0.117 & \textbf{0.145} & 0.044 & 0.120 & 0.119 & \textbf{0.148} \\
Cabécar & 0.044 & 0.125 & 0.127 & \textbf{0.157} & 0.041 & 0.119 & 0.121 & \textbf{0.147} \\
Cashinahua & 0.055 & 0.117 & 0.116 & \textbf{0.153} & 0.047 & 0.117 & 0.126 & \textbf{0.153} \\
\end{longtable}
\endgroup
\begingroup
\scriptsize
\setlength{\tabcolsep}{3pt}
\renewcommand{\arraystretch}{0.92}
\begin{longtable}{lcccc}
\caption{Language-wise multilingual AAS scores on DoReCo. Lower values are better. * Trained in this work.}\label{tab:aas_table_doreco}\\
\toprule
\textbf{Language} 
& \textbf{Base}& \textbf{MMS-DORECO*}& \textbf{MMS-DORECO*-SIL}& \textbf{Gold}\\
\midrule
\endfirsthead
\toprule
\textbf{Language} 
& \textbf{Base}& \textbf{MMS-DORECO*}& \textbf{MMS-DORECO*-SIL}& \textbf{Gold}\\
\midrule
\endhead
\bottomrule
\endfoot
Anal & 143.4 & 81.8 & 79.4 & \textbf{0.0} \\
Yali (Apahapsili) & 91.6 & 63.6 & 62.8 & \textbf{0.0} \\
Arapaho & 529.8 & 298.2 & 250.1 & \textbf{0.0} \\
Baïnounk Gubëeher & 172.3 & 101.9 & 92.4 & \textbf{0.0} \\
Beja & 57.7 & 52.8 & 53.1 & \textbf{0.0} \\
Bora & 426.1 & 178.0 & 171.2 & \textbf{0.0} \\
Cabécar & 230.7 & 111.7 & 113.5 & \textbf{0.0} \\
Cashinahua & 308.7 & 122.7 & 107.9 & \textbf{0.0} \\
\end{longtable}
\endgroup
\begingroup
\scriptsize
\setlength{\tabcolsep}{2.5pt}
\begin{longtable}{lp{2.8cm}cccccccc}
\caption{Multilingual statistics across aligners and embedding models. * Trained in this work.}\label{tab:multilingual-stats}\\
\toprule
\multirow{2}{*}{\textbf{Aligner}} & \multirow{2}{*}{\textbf{Metric}} & \multicolumn{4}{c}{\textbf{MMS}} & \multicolumn{4}{c}{\textbf{XLSR}} \\
\cmidrule(lr){3-6}\cmidrule(lr){7-10}
& & \textbf{Mean} & \textbf{Std} & \textbf{Min} & \textbf{Max} & \textbf{Mean} & \textbf{Std} & \textbf{Min} & \textbf{Max} \\
\midrule
\endfirsthead
\toprule
\multirow{2}{*}{\textbf{Aligner}} & \multirow{2}{*}{\textbf{Metric}} & \multicolumn{4}{c}{\textbf{MMS}} & \multicolumn{4}{c}{\textbf{XLSR}} \\
\cmidrule(lr){3-6}\cmidrule(lr){7-10}
& & \textbf{Mean} & \textbf{Std} & \textbf{Min} & \textbf{Max} & \textbf{Mean} & \textbf{Std} & \textbf{Min} & \textbf{Max} \\
\midrule
\endhead
\midrule
\endfoot
\bottomrule
\endlastfoot
Baseline (FLEURS) & PCMI & 0.091 & 0.017 & 0.048 & 0.148 & 0.093 & 0.016 & 0.053 & 0.148 \\
 & Mean cluster purity & 0.135 & 0.022 & 0.083 & 0.183 & 0.138 & 0.026 & 0.081 & 0.194 \\
 & Std cluster purity & 0.140 & 0.041 & 0.036 & 0.226 & 0.146 & 0.049 & 0.047 & 0.270 \\
 & Cluster entropy & 3.834 & 0.024 & 3.750 & 3.875 & 3.853 & 0.014 & 3.811 & 3.887 \\
 & Norm. cluster entropy & 0.980 & 0.006 & 0.959 & 0.991 & 0.985 & 0.004 & 0.974 & 0.994 \\
 & Label entropy & 3.501 & 0.206 & 3.047 & 3.946 & 3.501 & 0.206 & 3.047 & 3.946 \\
 & Norm. label entropy & 0.893 & 0.024 & 0.839 & 0.963 & 0.893 & 0.024 & 0.839 & 0.963 \\
 & Clusters used & 50.000 & 0.000 & 50.000 & 50.000 & 50.000 & 0.000 & 50.000 & 50.000 \\
\end{longtable}
\endgroup

\end{document}